\documentclass[twoside,11pt]{article}

\usepackage{blindtext}
\usepackage[preprint]{jmlr2e}
\usepackage{amsmath,amsfonts,amssymb}
\usepackage{bbm}
\usepackage{enumitem} 
\usepackage{floatrow} 
\usepackage{subcaption} 
\usepackage{xcolor}
\usepackage{breqn} 
\usepackage{rotating} 
\usepackage{soul} 

\usepackage{adjustbox}
\usepackage{longtable}
\usepackage{tabularx}
\usepackage{tablefootnote}
\usepackage{multirow}

\usepackage{booktabs} 

\newcommand{\cmmnt}[1]{}

\usepackage{easy-todo}
\usepackage{cancel}
\usepackage{ulem}

\usepackage{tikz}
\usetikzlibrary{positioning}
\usetikzlibrary{trees}
\usetikzlibrary{shapes.geometric, arrows}

\DeclareMathOperator*{\argmax}{argmax}

\usepackage{lastpage}
\jmlrheading{}{2025}{1-\pageref{LastPage}}{}{}{}{Christos Revelas, Otilia Boldea and Bas J.M. Werker}
\ShortHeadings{Consistency of Selection Strategies for Fraud Detection}{Revelas, Boldea and Werker}
\firstpageno{1}

\begin{document}

\title{Consistency of Selection Strategies for Fraud Detection}


\author{\name Christos Revelas \email c.revelas@tilburguniversity.edu\\
	\name Otilia Boldea \email o.boldea@tilburguniversity.edu\\
	\name Bas J.M. Werker \email b.j.m.werker@tilburguniversity.edu\\
       \addr Tilburg University \\
       Department of Econometrics and Operations Research\\
       Warandelaan 2, 5037 AB Tilburg, Netherlands
      }


\maketitle

\begin{abstract}
This paper studies how insurers can chose which claims to investigate for fraud. Given a prediction model, typically only claims with the highest predicted propability of being fraudulent are investigated. We argue that this can lead to inconsistent learning and propose a randomized alternative. More generally, we draw a parallel with the multi-arm bandit literature and argue that, in the presence of selection, the obtained observations are not iid. Hence, dependence on past observations should be accounted for when updating parameter estimates. We formalize selection in a binary regression framework and show that model updating and maximum-likelihood estimation can be implemented as if claims were investigated at random. 
Then, we define consistency of selection strategies and conjecture sufficient conditions for consistency. Our simulations suggest that the often-used selection strategy can be inconsistent while the proposed randomized alternative is consistent. Finally, we compare our randomized selection strategy with Thompson sampling, a standard multi-arm bandit heuristic. Our simulations suggest that the latter can be inefficient in learning low fraud probabilities. \\
\end{abstract}


%


\section{Introduction}
Fraud is a common phenomenon in insurance and detecting fraud can be beneficial for insurance companies and their customers. However, detecting fraud comes with costs as investigating claims uses human resources and can negatively impact customer satisfaction. Choosing which claims to investigate for fraud is therefore an important question that insurers face in their day-to-day business. 

Traditionally insurers have flagged claims based on a set of expert rules: if a claim triggers one or more of these rules, then it is further examined and, if deemed worth it, investigated. Only then fraud can be established with high certainty. This creates selection bias: fraud is only observed for claims that have been investigated. With the incorporation of statistical modelling and predictive analytics in businesses in the past decades, insurers started modelling fraud either to replace the traditional expert-based approach or to complement it. 

The existing literature on fraud detection is large and covers several aspects, ranging from ways to improve the already in-place rule-based detection systems (e.g., \citealp{baumann2021improving} look for correlations between existing rules, and \citealp{liu2020automobile} put weights in existing rules by combining them with historical data), to data-engineering techniques (\citealp{baesens2021data}) and resampling methods that address class imbalance, or skewness, a typical characteristic of fraud datasets (see, e.g., \citealp{baesens2021robrose}). Selection bias has been adressed for example in \citealp{pinquet2007selection}, who propose a bias correction method in a dataset where some of the claims were investigated at random; \citealp{caudill2005fraud} propose a logit model that accounts for misclassification error; \citealp{barton2024importance}, in the context of accounting fraud, propose a bivariate probit model to model separately investigation and fraud. Cost-sensitive models, i.e., when the cost of misclassifying a fraudulent claim as non-fraudulent is different from the cost of misclassifying a non-fraudulent claim as fraudulent, have been proposed for example in \citealp{hoppner2022instance}. Fraud detection can be studied as an unsupervised anomaly detection problem, see, e.g., \citealp{lu2006adaptive}, \citealp{stripling2018isolation}, \citealp{snorovikhina2021unsupervised} and \citealp{cong2023crypto}. In supervised learning, many classical learning methods have been applied, such as logistic regression, support vector machines (e.g. \citealp{cecchini2010detecting} in management fraud) and ensemble methods; see, e.g., \citealp{lessmann2008benchmarking} for a benchmark of several methods in the context of software defect prediction, or \citealp{schrijver2024automobile} and \citealp{benedek2022automobile} for recent literature reviews of fraud detection in automobile car insurance in particular. Among studies that apply deep learning and artificial intelligence methods we can cite, e.g., \citealp{viaene2005auto} who apply Bayesian neural networks in car insurance fraud detection, \citealp{van2017gotcha} and \citealp{oskarsdottir2022social}, who establish fraud as a social phenomenon and estimate network relationships between policy holders, insurance brokers and repair garages, or \citealp{dimri2022multi} who apply large language models to text associated with claims. 

The present paper looks at fraud detection as a supervised learning problem, and, rather than focusing on fraud data characteristics and methodologies to adress it, we are interested in how a given (supervised) prediction model can be updated with the arrival of new claims (see, e.g., the deployment perspective of \citealp{baesens2023fraud}). Once a prediction model in place, it is common for insurers to use this model for selection of claims to be investigated in a deterministic manner: claims with the highest predicted fraud probability are always selected for investigation, while claims at the bottom of predictions are never chosen. This poses a concern for self-selection which can lead to inconsistent learning and, consequently, suboptimal fraud detection. We point out that, when selecting claims for investigation, there is always a trade-off between detecting fraud, in other words exploitation of past knowledge, and learning about fraud, i.e., exploration for new types of fraud (see, e.g., \citealp{soemers2018adapting}, who point out the same trade-off in the context of credit card fraud detection). Hence, the common strategy of always choosing claims with the highest predicted fraud probability might not be the best practice. We propose an alternative strategy which addresses the exploration-exploitation trade-off: we randomize selection such that, claims with the highest predicted fraud probability are \textit{most-often} selected for investigation but not always, while claims at the bottom of predictions are chosen rarely but with positive probability.

The exploration-exploitation trade-off has been studied in the multi-arm bandit (MAB) literature where, typically, the goal is to maximize a cumulative reward. In the classical MAB problem (see, e.g., \citealp{auer2002finite}), an agent chooses among arms to ``play" at each point in time based on prior beliefs on each arm's reward probabilities. One would want to play the bandit with the highest payoff, but that entails estimation error and, therefore, may get ``stuck" playing the suboptimal arm. The fraud detection problem can be formulated as a MAB problem by dividing the covariate or claim characteristics space into groups (or arms) and then choosing claims for investigation from the different groups, assuming a constant fraud probability within each group. Alternatively, the fraud detection problem can be formulated as a contextual MAB (CMAB, see, e.g., \citealp{10.1214/13-AOS1101}) problem, where, additionally, a functional relation between covariates and target variable can be defined. See, e.g., \citet{soemers2018adapting} for an application of CMABs to credit card fraud detection. Other examples include \citealp{collier2018deep}, who apply CMABs in a marketing context, and \citealp{jung2012contextual} in the context of spam prevention. For a survey of other applications of (C)MABs see, e.g., \citealp{bouneffouf2019survey}. 
While in the MAB literature the goal typically is to maximize a cumulative reward, these methods can also be applied with a focus on learning, which is the perspective we take.

The present paper contributes in four ways to the above mentioned literature. 
First, we argue that in (C)MABs as above, model updating should account for the fact that the sequence of claims pulled for investigation does not form an iid sample: which arm is pulled at any given time depends on all past chosen arms. 
Second, we formalize non-iid selection. We do this in a binary regression model with Bayesian updating, but our findings extend to the (C)MAB literature when the relation between contexts and reward is also assumed to follow a binary regression model. In our framework, we define selection strategies as maps from posteriors for the parameter in the binary regression to distributions on the covariate (or feature) space. We show that, if selection is independent of the parameter of the regression, then model updating and maximum likelihood estimation can both be correctly implemented as if observations where iid, which is of practical significance. 
Third, we define consistency of selection strategies arguing that this is a desirable property for any fraud detection process implemented in practice. We draw from the literature on Bayesian and maximum-likelihood estimation to conjecture sufficient conditions for consistency. In particular, we compare two selection strategies. On the one hand, the typical strategy where the insurer picks claims with the highest predicted fraud probability as described earlier. On the other hand, a randomized alternative where the insurer draws at random claims proportionally to their predicted fraud probability. We present simulation results which suggest that the typical selection mechanism used by insurers might not be consistent while the proposed randomized analogue appears to correctly learn the model's parameters. This suggests that, in practice, randomization can lead to better prediction models over time.
Fourth and finally, we compare our randomized strategy to Thompson sampling (see, e.g., \citealp{thompson1933likelihood}, \citealp{agrawal2012analysis}, \citealp{kaptein2015use}). Thompson sampling is a MAB strategy which constructs a posterior distribution for the reward (in this case, fraud detection) on each arm (assuming previously observed rewards for that arm are iid), draws a potential reward for each arm from those posteriors and then selects the arm with the highest drawn reward. This strategy differs from our randomized proposal in that it does not assume any global parametric form for the fraud probability, which is assumed constant within each arm. When interpretability is important, this can be a disadvantage. Our strategy on the other hand resembles Thompson sampling in that both draw randomly claims proportionally to their believed fraud likelihoods. We present a simulation comparison between these two strategies which suggests that, in the presence of a global parametric model, Thompson sampling can be inefficient in regions of low fraud probability. 

The remainder of this paper is organized as follows. Section \ref{section_framework} sets the mathematical framework. We formalize selection, show that Bayes' formula yields the same sequence of posteriors as when (possibly incorrectly) assuming that claims are investigated at random. Note that the probabilistic behavior of the sequence of posteriors is different. We also give examples of selection strategies. In Section \ref{section_consistency} we define consistency of selection strategies and discuss consistency. Then we compare in simulations the most-likely and randomized strategies. In Section \ref{section_thompson} we compare the randomized strategy to Thompson sampling. Section \ref{section_practical_guidelines} contains practical considerations. Section \ref{section_conclusion} concludes. Proofs are gathered in the appendix.


\section{Selection strategies for fraud detection}\label{section_framework}
Here we set the mathematical framework for the remainder of the paper. We consider a parametric binary regression to model insurance fraud. We formalize the selection of claims and show that, under some assumptions, model updating can be implemented as if there was no selection. Finally, we present and discuss particular examples of selection strategies. 

\subsection{Bayesian updating for non-iid data}

We assume that fraud follows a binary regression model of the form
\begin{equation}
Y|X,\theta\sim{\textrm Bernoulli}(g(X'\theta))
\label{model}
\end{equation}
where $Y\in\{0,1\}$ and $Y=1$ represents fraud, $X\in\mathbb{R}^k$ corresponds to $k$ claim characteristics, and $\theta\in\Theta\subset\mathbb{R}^k$ weights each characteristic to the likelihood of a claim being fraudulent. 
We assume throughout that $X$ is continuous and of known density $p_X$. We assume that the function $g:\mathbb{R}\rightarrow[0,1]$ is continuous and strictly monotone. For example, $g(u)=(1+\exp(-u))^{-1}$ corresponds to the logit model. We consider $\theta$ to be a random variable drawn before everything else by nature from some unknown distribution\footnote{ Throughout we use the notation $\theta$ for the random variable and $\vartheta$ for local values, e.g., in integrals.}. We formalize it in this way to be able to use martingale convergence results for consistency in Section \ref{section_consistency}. Finally, we assume given a starting prior distribution for $\theta$ of density $\pi_0$ on $\Theta$. In practice, $\pi_0$ can be chosen arbitrarily. 

We consider throughout this paper samples $(X_1,Y_1),\dots,(X_n,Y_n)$ of observations that are not necessarily iid conditionally on $\theta$ but satisfy, for all $1\leq i\leq n$,
\begin{equation}
p(Y_{i+1}|X_1,Y_1,\dots,X_i,Y_i, X_{i+1}, \theta) = p(Y_{i+1}|X_{i+1}, \theta)
\label{dgp_assumption}
\end{equation}
where $p$ denotes probability distributions, and $p(Y_{i+1}|X_{i+1}, \theta)$ is given by (\ref{model}). In other words whether a given claim is fraudulent or not does not depend on past claims. 

We wish to allow for the practitioner to select which claims to investigate for fraud. We formalize this selection mechanism, or selection strategy, as a map from past observations $x_1,y_1, \dots, x_i,y_i$ to a density on the covariate space, corresponding to the density of $X_{i+1}$ and from which $X_{i+1}$ is then drawn. Once $X_{i+1}$ is selected, $Y_{i+1}$ is then drawn from (\ref{dgp_assumption}). Further, we assume that selection does not depend on the parameter $\theta$ conditionally on past observations, a key assumption for model updating. In line with classical Bayesian ideas, the researcher summarizes the knowledge obtained at stage $i$ by a density $\pi_i$ on $\Theta$. Then the researcher uses Bayes rule to obtain $\pi_{i+1}$ from $\pi_i$ and the observed values $X_{i+1} = x_{i+1}$ and $Y_{i+1} = y_{i+1}$. Finally, in the present paper, we consider selection mechanisms in which the researcher only looks at the past through the density $\pi_i$ at any given stage $i$, i.e., we assume that
\begin{equation} 
p(x_{i+1}|x_1,y_1,\dots,x_i,y_i) = p(x_{i+1}|\pi_i). 
\label{selection_assumption_2} 
\end{equation}
This assumption is sufficient to cover the selection mechanisms considered in the remainder of the paper, but is not necessary for our theory.  

Formally, let $\Pi$ and $\Delta_X$ be the sets of distributions on $\Theta$ and the covariate space $\mathbb{R}^k$ respectively\footnote{ For simplicity, here we omit the formal treatment of distributions as maps on measurable spaces and only use this formalism in later proofs.}.
\begin{definition}\label{definition_strategy}
A \textit{selection strategy} is a map from past observations to $\Delta_X$ such that, for every $i$,
\begin{equation} 
p(x_{i+1}|x_1,y_1,\dots,x_i,y_i,\vartheta) = p(x_{i+1}|x_1,y_1,\dots,x_i,y_i).
\label{selection_assumption} 
\end{equation}
When, in addition, (\ref{selection_assumption_2}) is satisfied, i.e., when the researcher only looks at the past through the posterior, then the selection strategy is reduced to a map from $\Pi$ to $\Delta_X$.
\end{definition}
Note that a selection strategy is a deterministic map. Moreover, we need to assume that a claim for each possible value of the map's image distribution is available. 

Next we show that, when properly taking into account the selection mechanism when applying Bayes' rule, the selection drops out of the equation.
\begin{proposition}\label{proposition_bayes}
For every data-generating process satisfying (\ref{model}) and (\ref{dgp_assumption}) and every selection strategy satisfying (\ref{selection_assumption}), noting $\pi_{i}$ and $\pi_{i+1}$ the densities $p(\vartheta|x_1,y_1,\dots,x_i,y_i)$ and $p(\vartheta|x_1,y_1,\dots,x_i,y_i,x_{i+1},y_{i+1})$ respectively, we have, for every $i$,
\begin{equation}
\pi_{i+1} = \frac{p(y_{i+1}|x_{i+1},\vartheta)\pi_i}{p(y_{i+1}|x_{i+1})}.
\label{bayes_1}
\end{equation}
\end{proposition}
A proof is given in Appendix \ref{appendix_bayes}. In other words, we may use Bayes' rule as if the selection mechanism was using the same density at each stage, independently of past observations, thus as if the sample $(X_1,Y_1),\dots,(X_n,Y_n)$ was iid given $\theta$. Proposition \ref{proposition_bayes} holds in particular for posterior-based strategies. We stress that the probabilistic behavior of the densities $\pi_i$ does depend on the dependence in consecutive observations $(X_i,Y_i)$.

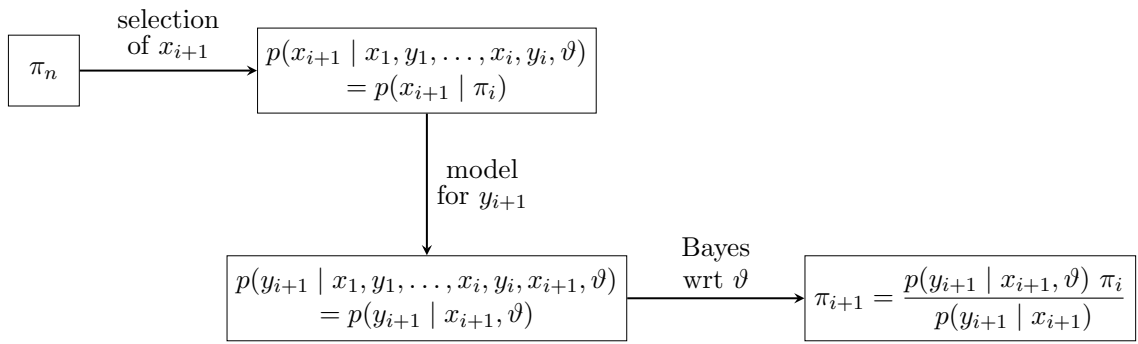
\begin{figure}[htbp]
\centering
\resizebox{\textwidth}{!}{ 
\begin{tikzpicture}[
    node distance=2cm and 2.5cm, 
    block/.style={rectangle, draw=black, fill=white, minimum width=4.5cm, minimum height=1.2cm, align=center},
    smallblock/.style={rectangle, draw=black, fill=white, minimum width=1cm, minimum height=1cm, align=center},
    arrow/.style={thick,->,>=stealth}
]
\node (p) [smallblock] {$\pi_n$};
\node (X1) [right=of p, block] {\shortstack{$p(x_{i+1} \mid x_1,y_1,\dots,x_i,y_i,\vartheta)$\\$=p(x_{i+1} \mid \pi_i)$}}; 
\node (Y1) [below=of X1, block] {\shortstack{$p(y_{i+1} \mid x_1,y_1,\dots,x_i,y_i,x_{i+1},\vartheta)$\\$=p(y_{i+1} \mid x_{i+1}, \vartheta)$}};
\node (p1) [right=of Y1, block] {$\pi_{i+1}=\displaystyle\frac{p(y_{i+1} \mid x_{i+1},\vartheta) \ \pi_i}{p(y_{i+1} \mid x_{i+1})}$};
\draw [arrow] (p) -- node[above] {\shortstack{selection \\ of $x_{i+1}$}} (X1);
\draw [arrow] (X1) -- node[right] {\shortstack{model \\ for $y_{i+1}$}} (Y1);
\draw [arrow] (Y1) -- node[above] {\shortstack{Bayes \\ wrt $\vartheta$}} (p1);
\end{tikzpicture}
}
\caption{Selection and updating}
\label{selection_and_update}
\end{figure}

We summarize selection and updating in Figure \ref{selection_and_update}.

\subsection{Example strategies and the exploration-exploitation trade-off}
The above formalism allows to encompass several selection mechanisms. For example, a naive strategy would be to always choose claims with the same characteristics, independently of the posterior belief at each step. Formally, this strategy is a map of the form 
$\pi_i\mapsto \delta_{a}$
for all $i$, where $\delta$ denotes the degenerate distribution at some fixed value $a\in\mathbb{R}^k$. 

An insurer typically selects claims based on their predicted likelihood of being fraudulent. Given a prediction model, claims are ordered and those with the highest predicted fraud probability are selected for investigation. We call this the most-likely strategy and formalize it as follows within our framework. Let for all $i$
\begin{equation}
\theta^{\pi}_i:=\int_{\vartheta\in\Theta}\vartheta \pi_i(\vartheta) d\vartheta
\label{definition_posterior_mean}
\end{equation} 
be the $i^{th}$-posterior mean of $\theta$. 
\begin{definition}
The \textit{most-likely selection} strategy is the map 
\begin{equation}
\pi_i\mapsto \delta_{\argmax_{x\in\mathbb{R}^k}g(x'\theta^{\pi}_i)}
\end{equation}
where $\delta_{\argmax_{x\in\mathbb{R}^k}g(x'\theta^{\pi}_i)}$ is the degenerate distribution at the covariate value that maximizes the believed fraud likelihood based on $\theta^{\pi}_i$.
\end{definition}
While this is a natural selection strategy in a reward-maximization setting such as insurance fraud detection, in the present paper we point out and later show that this strategy can lead to inconsistent learning. When selecting claims for investigation, there is an inherent trade-off between detecting fraud and learning about fraud. The most-likely selection solely focuses on the former. To the other extreme, one could focus solely on learning, e.g., by selecting claims at random independently of the posterior at each step. We call this the iid strategy because the obtained sample $(X_1,Y_1,\dots,X_n,Y_n)$ is iid conditionally on $\theta$.
\begin{definition}
The \textit{iid selection} strategy is the map
$\pi_i\mapsto p_X$,
where $p_X$ is the distribution of $X$.
\end{definition}
This strategy does not take into account the posterior beliefs and, particularly in the presence of class imbalance, will tend to perform poorly in detecting fraud. In the present paper we propose a selection mechanism that lies somewhere between the most-likely and iid cases. We call it the randomized most-likely and its rationale is to still prioritize claims with high predicted fraud probabilities while allowing for some exploration. 
\begin{definition}\label{definition_RML_strategy}
Define, for every $i$, a distribution $p_i$ on $\mathbb{R}^k$ by its density
\begin{equation}
p_i(x) := \frac{g(x'\theta^{\pi}_i)}{\int_{\xi\in\mathbb{R}^k}g(\xi'\theta^{\pi}_i)p_X(d\xi)}
\label{prob_weights_RML}
\end{equation}
where $\theta^{\pi}_i$ is as in (\ref{definition_posterior_mean}) and we recall that $X$ is assumed continuous of density $p_X$, with $p_X(d\xi)=p_X(\xi)d\xi$. The \textit{randomized most-likely selection} strategy is the map
$\pi_i\mapsto p_i$.
\end{definition}
In other words, we select claims at random but proportionally to their believed fraud likelihood based on $\theta^{\pi}_i$: the more likely a claim is believed to be fraudulent, the more likely it is to be selected. 



\section{Consistency of selection strategies}\label{section_consistency}
We are interested in consistent learning, i.e., over time, correctly learning $\theta$. In this section we define consistency of selection strategies and give sufficient conditions for a strategy to be consistent. First, we establish a general sufficient condition for consistency drawing from Bayesian literature that uses martingale theory (see, e.g., \citealp{miller2018detailed}, \citealp{van2000asymptotic}, \citealp{doob1949application}). Consistency can be proven in two steps: (i) by showing that the sequence of posteriors means converges; and (ii) by showing that the parameter can be measurably recovered from observations. The posterior convergence holds for any selection strategy because the sequence of posterior means forms a martingale. Hence, recovery implies consistency (Theorem \ref{theorem_consistency}). The recovery condition however typically assumes iid data. Second, we show that under our selection assumption (\ref{selection_assumption}), the maximum likelihood estimator is the same as in the iid case. Arguing that the existence of a strongly consistent sequence of estimators for $\theta$ implies the recovery condition, we then conjecture a sufficient condition for consistency of the maximum likelihood estimator (Conjecture \ref{conjecture_empirical}) and, hence, for consistency of a selection strategy.

Throughout this section, we note again $\{\pi_n\}_{n\geq1}$ and $(X_1,Y_1),\dots,(X_n,Y_n)$ respectively the sequence of posteriors and sample of observations obtained from a selection strategy as in Definition \ref{definition_strategy}. Intuitively, consistency is obtained when the posteriors concentrate more and more around the parameter. Formally, we define consistency as the weak convergence of the sequence of posteriors to a distribution that is degenerated at $\theta$.
\begin{definition}\label{definition_consistency}
A selection strategy is \textit{consistent} if for every continuous and bounded function $h:\Theta\longrightarrow\mathbb{R}$ we have, as $n\rightarrow\infty$, almost surely
\begin{equation}
\int_{\vartheta\in\Theta}h(\vartheta)\pi_n(\vartheta)d\vartheta \rightarrow h(\theta) 
\end{equation}
where $\theta$ is the true parameter\footnote{ Note that from the boundness of $h$, $\int_{\vartheta\in\Theta}h(\vartheta)\pi_n(\vartheta)d\vartheta$ exists as $\pi_n$ integrates to 1. }. 
\end{definition}
We note $(X_1,Y_1),\dots,(X_\infty,Y_\infty)$ the limit as $n\rightarrow\infty$ of the sample $(X_1,Y_1),\dots,(X_n,Y_n)$ obtained from a selection strategy. 
\begin{theorem}\label{theorem_consistency}
A selection strategy is consistent if there exists a measurable function
$Q:(\mathbb{R}^k\times\{0,1\})^\infty \rightarrow \Theta$ such that, almost surely,
\begin{equation}
\theta=Q\big((X_1,Y_1),\dots,(X_\infty,Y_\infty)\big).
\label{recovery_condition}
\end{equation}
\end{theorem}
A proof is given in Appendix \ref{appendix_consistency_thm}. Condition (\ref{recovery_condition}) says that we can recover the parameter from observations in a measurable way. Theorem \ref{theorem_consistency} holds for any selection strategy as defined in Definition \ref{definition_strategy} but this does not mean that any selection strategy is consistent. The recovery condition may or may not hold for a given strategy. Verifying whether (\ref{recovery_condition}) holds for a given selection mechanism is not trivial. One way to verify (\ref{recovery_condition}) is through maximum likelihood, which is the focus of the next subsection.

\subsection{Maximum likelihood for non-iid data}

We start with the following observation.
\begin{proposition}\label{proposition_MLE}
For any strategy as in Definition \ref{definition_strategy}, the maximum log-likelihood estimator is given by
\begin{equation} 
\hat{\vartheta}_n=\argmax_{\vartheta\in\Theta}\sum_{i=1}^n\log P(Y_i|X_i,\vartheta).
\end{equation}
\end{proposition}
Proposition \ref{proposition_MLE} is in general not trivial because, on the one hand, selection might depend on $\theta$, in which case the likelihood of $X_i$ (given the past) does not simplify and, on the other hand, the likelihood of $Y_i$ might depend on the past. A proof is given in Appendix \ref{appendix_proposition_likelihood}. As in Proposition \ref{proposition_bayes}, assuming that selection does not depend on $\theta$ conditionally on the past is key. Proposition \ref{proposition_MLE} says that we get the same estimator as if there was no selection, i.e., as in the standard case with iid data. Again, this does not imply that the probabilistic behavior of the estimator is the same as under iid data. Then, we claim the following. Let $\mathbf{X}:=(X_{ij})_{1\leq i\leq n, 1\leq j\leq k}$ be the design matrix obtained from a selection strategy, where $X_{ij}$ denotes the $j^{th}$ coordinate of $X_i$. 
\begin{conjecture}\label{conjecture_empirical}
Under model (\ref{model}) and its assumptions, a selection strategy is consistent if the limit as $n\rightarrow\infty$ of 
\begin{equation}
\frac{1}{n}\mathbf{X}'\mathbf{X} = \frac{1}{n}\sum_{i=1}^n X_i X_i'
\label{sufficient_condition_limit}
\end{equation}
exists, is non-random and invertible.
\end{conjecture}
We give some suggestions towards a possible proof in Appendix \ref{appendix_conjecture}. Conjecture \ref{conjecture_empirical} is not trivial because of non-iid data. For consistency to hold, we typically need a uniform law of large numbers to hold, which in turn requires that observations are sufficiently independent. We believe that, for the limit $\frac{1}{n}\mathbf{X}'\mathbf{X}$ to exist, we also need sufficiently independent observations. Moreover, as our simulations also suggest, the sufficient condition of Conjecture \ref{conjecture_empirical} seems to also be necessary.

\subsection{Examples}
Here we examine (\ref{sufficient_condition_limit}) in particular examples in two dimensions ($k=2$). 

\begin{example}[naive]
Let $a=(1,1)'$. The naive strategy $\pi_i\mapsto \delta_{a}$ always selects claim $(1,1)'$ irrespective of the posterior and the DGP distribution $p_X$. Then, for all $n$,
\begin{equation}
\frac{1}{n}\sum_{i=1}^n X_i X_i' = \begin{pmatrix} 1 & 1\\ 1 & 1\end{pmatrix}
\label{example_singular}
\end{equation}
is of rank 1 and hence the limit is not invertible.
\end{example}

\begin{example}[iid] 
Suppose that $X$ in (\ref{model}) follows a bivariate normal distribution $p_X\sim\mathcal{N}(\mu,\Sigma)$ with $\mu=(\mu_1,\mu_2)'$ and $\Sigma=\begin{pmatrix} \sigma_1^2 & \sigma_{12}\\ \sigma_{12} & \sigma_2^2 \end{pmatrix}$. The iid strategy selects claims from the DGP, i.e., $\pi_i\mapsto p_X$. In this case, we can apply a standard LLN with iid data to obtain the limit
\begin{equation}
\frac{1}{n}\sum_{i=1}^n X_i X_i' \rightarrow \mathbb{E}[XX']=\begin{pmatrix} \mu_1+\sigma_1^2 & \mu_1\mu_2+\sigma_{12}\\ \mu_1\mu_2+\sigma_{12} & \mu_2+\sigma_2^2 \end{pmatrix}
\end{equation}
which is invertible iff $(\mu_1+\sigma_1^2)(\mu_2+\sigma_2^2)-(\mu_1\mu_2+\sigma_{12})^2\neq0$. 
\end{example}

Next we assume that the function $g$ in (\ref{model}) is the logistic function. Moreover, we assume a normal starting prior $\pi_0\sim\mathcal{N}(\mu_0,\Sigma_0)$ with $\mu_0=(2,1)'$ and $\Sigma=\mathrm{diag}(0.75,0.75)$.

\begin{example}[most-likely]
The most-likely strategy selects claims $x$ maximizing $g(x'\theta^{\pi}_i)$ where $\theta^{\pi}_i$ is the posterior mean as defined in (\ref{definition_posterior_mean}). In simulations we observe that this strategy always selects claim $(1,1)'$, hence behaving as the naive strategy example and again the limit (\ref{example_singular}) is not invertible. 
\end{example}

\begin{example}[randomized most-likely]
The randomized most-likely strategy selects claims at random drawing from the distribution $p_i(x) = \frac{g(x'\theta^{\pi}_i)}{\int_{\xi\in\mathbb{R}^k}g(\xi'\theta^{\pi}_i)dp_X(\xi)}$.  In simulations we observe that this strategy selects claims of varying values around (0.5,0.5), and that $\frac{1}{n}\sum_{i=1}^n X_i X_i'$ is invertible for large $n$.
\end{example}


\subsection{Simulations}\label{subsection_simulations}

The purpose of this section is to illustrate via simulations that the most-likely selection strategy can in some cases be inconsistent while the randomized most-likely strategy is consistent. One possible issue associated with the most-likely strategy is that it might stop learning by always selecting completely uninformative claims, i.e., with zero Fisher information, in which case the likelihood is flat and the posteriors get ``stuck". A second possible issue is that this strategy might keep selecting the same claim over and over, which can have a flattening effect on the empirical likelihood in some directions, depending on how informative the claim is. In more than one dimensions, the above two issues can happen separately. Here, we give an example in two dimensions in which the most-likely selection does not stop learning but is inconsistent and the limit of (\ref{sufficient_condition_limit}) is degenerate. In the same example, the randomized most-likely strategy is consistent as it allows for variation in the choice of claims. Here we consider a 2-dimensional logit DGP with $X\in[0,1]^2$ and $\Theta=\mathbb{R}^2$. We take $\vartheta_{true}=(\vartheta_{true}^{(1)},\vartheta_{true}^{(2)})'=(1,1)'$ and $\pi_0\sim\mathcal{N}(\mu_0,\Sigma_0)$ with $\mu_0=(2,1)'$ and $\Sigma=\mathrm{diag}(0.75, 0.75)$
. We run 50 realizations of each selection strategy with sample size $n=1000$. 

\begin{figure}[htbp]
\begin{subfigure}{0.49\textwidth}
	\includegraphics[width=\textwidth]{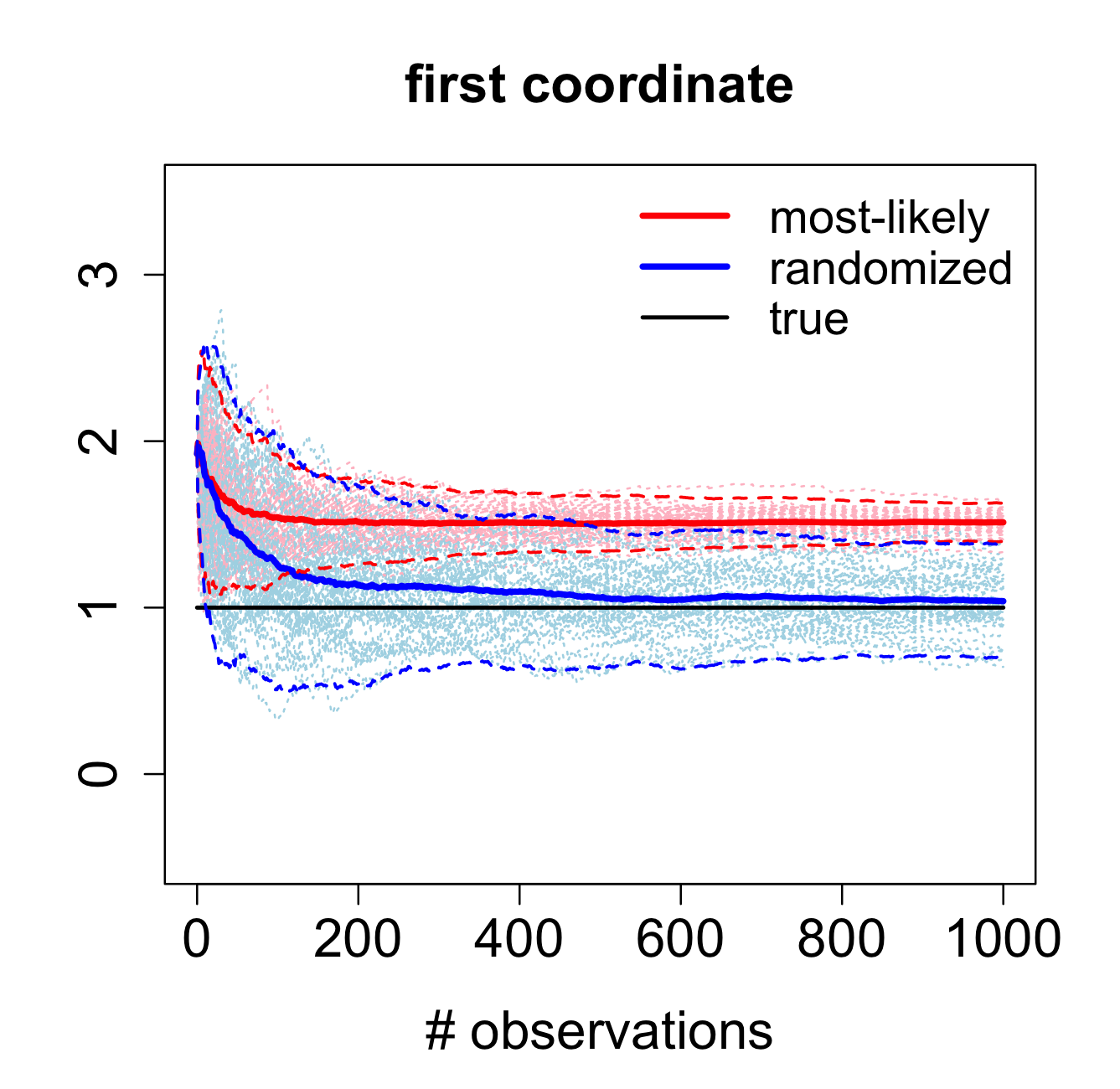}
\end{subfigure}
\begin{subfigure}{0.49\textwidth}
	\includegraphics[width=\textwidth]{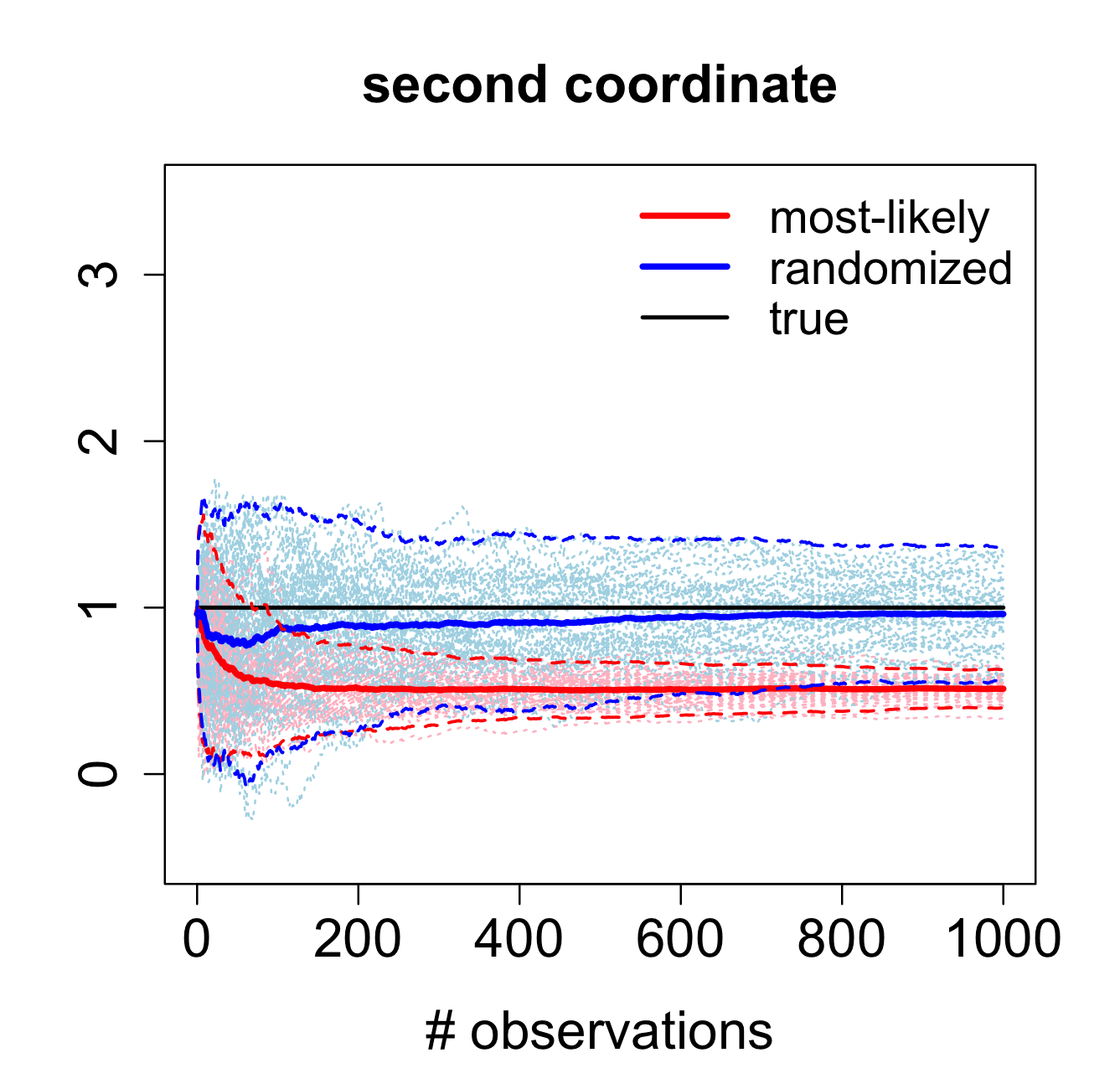}
\end{subfigure}
\caption{Estimates for each coordinate of the expected posterior mean $\theta^{\pi}_{n}$, as defined in (\ref{definition_posterior_mean}), for the most-likely strategy (red solid line) versus the randomized most-likely strategy (in blue). The dashed lines correspond to the average posterior mean $\pm$ two standard deviations over 50 realizations. The dotted lines are the 50 realizations. In black are the true values $\theta = (1,1)$. The most-likely strategy stabilizes around wrong values for each coordinated, suggesting inconsistency. The randomized strategy stabilizes around the true values.}
\label{2Destimates}
\end{figure}

Figure \ref{2Destimates} shows the posterior means $\theta^{\pi}_n=\int_{\vartheta\in\Theta}\vartheta \pi_i(\vartheta) d\vartheta$ for each strategy as a function of $n$. The left plot shows the first coordinate and the right plot shows the second coordinate. Dotted lines correspond to realizations, solid lines to their average and the dashed lines correspond to the average $\pm2$ standard deviations. In black are the true parameter values $\vartheta^{(1)}_{true}$ (left plot) and $\vartheta^{(2)}_{true}$ (right plot). We observe that the most-likely strategy always selects $X_i=(1,1)'$. This value being different from $(0,0)$, i.e. the only uninformative case in this example, the most-likely strategy will not stop updating. However, this strategy is inconsistent. The randomized most-likely strategy on the other hand shows variation in the selection, choosing claims around $(0.5,0.5)'$, and consistently estimates $\vartheta_{true}$. 

\begin{figure}[htbp]
\begin{subfigure}{0.49\textwidth}
	\includegraphics[width=\textwidth]{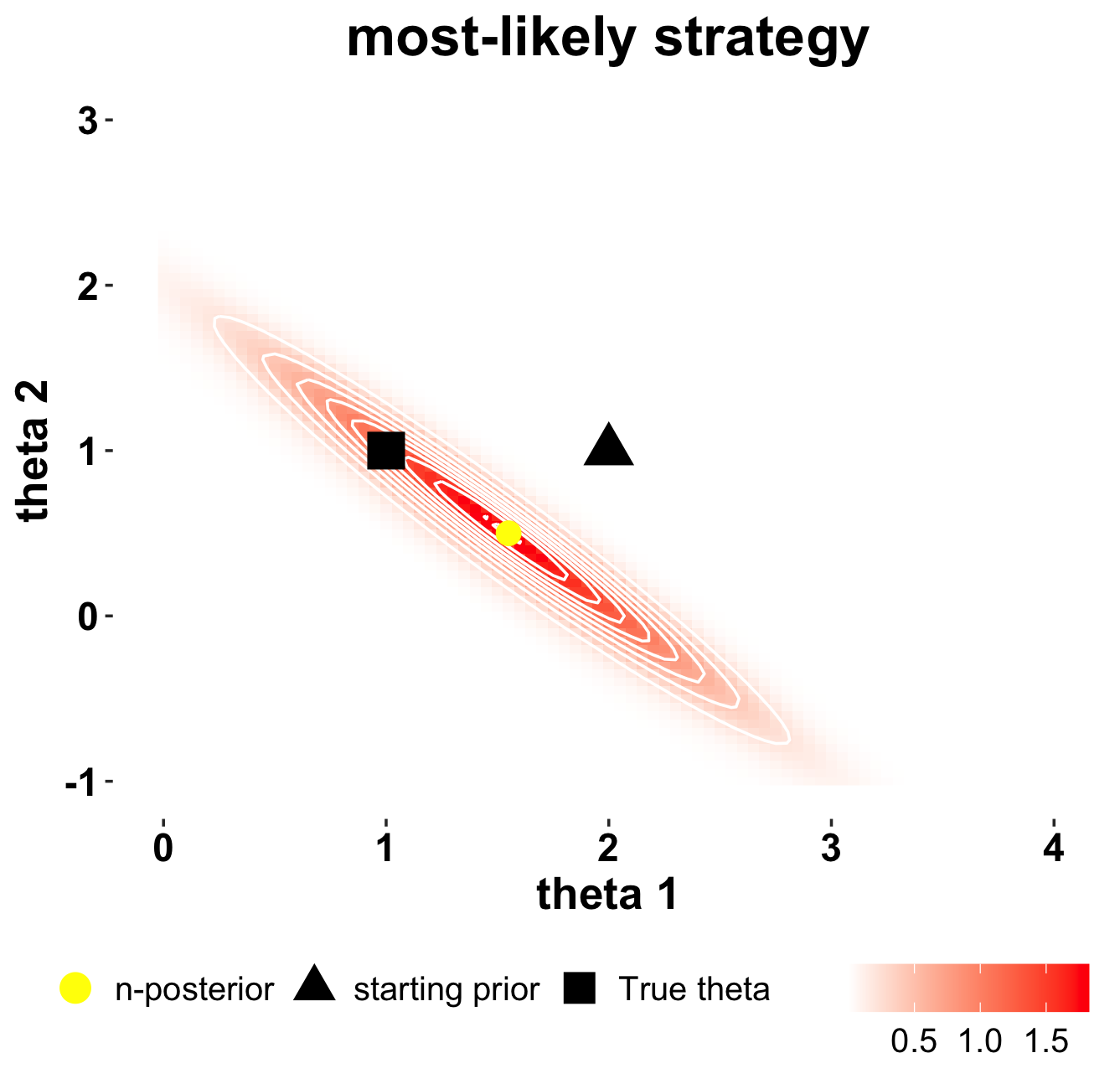}
\end{subfigure}
\begin{subfigure}{0.49\textwidth}
	\includegraphics[width=\textwidth]{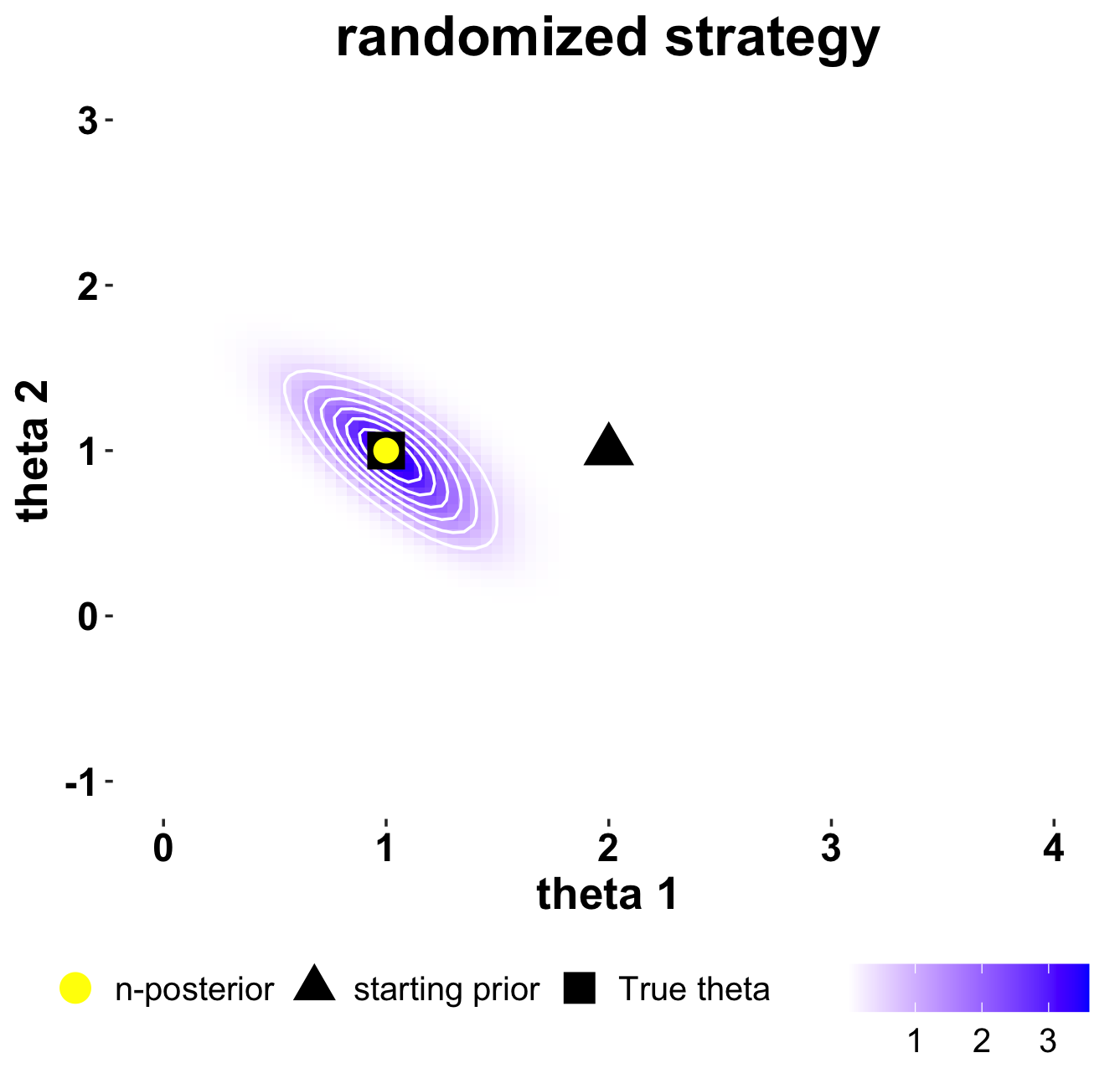}
\end{subfigure}
\caption{Estimates of the expected posterior $\pi_{1000}(\vartheta)$ for the most-likely (left) versus the randomized most-likely (right) strategies. Estimates are based on 50 replications. The color bar shows the posterior's level of concentration. The triangle and square correspond respectively to the starting prior mean $(2,1)$ and true parameter $\theta=(1,1)$. The yellow point is the average $1000^{th}$ posterior mean. For the most-likely strategy the posterior appears to not concentrate as it is stretched, and the posterior mean is far from the true value. In the case of the randomized strategy the posterior concentrates around the true value.}
\label{2Dposteriors}
\end{figure}

Figure \ref{2Dposteriors} shows the average (over realizations) terminal posterior $\pi_{1000}(\vartheta)$ as a function of $\vartheta$ for each strategy. The left plot shows the level of the average, over the realizations, terminal posterior $\pi_n(\vartheta)$ for the most-likely strategy, and the right plot for the randomized most-likely strategy. The yellow points correspond to the average (over realizations) posterior mean $\int_{\vartheta\in\Theta}\vartheta d\pi_n(\vartheta)$ of each strategy. The black square represents the true parameter. The triangle corresponds to the starting prior mean $\int_{\vartheta\in\Theta}\vartheta d\pi_0(\vartheta)$ which is the same for both strategies and deterministic. We see that the posterior is stretched in the case of the most-likely strategy: it does not seem to concentrate around $\vartheta_{true}$, while it does for the randomized most-likely strategy. 

\begin{figure}[htbp]
\begin{subfigure}{0.49\textwidth}
	\includegraphics[width=\textwidth]{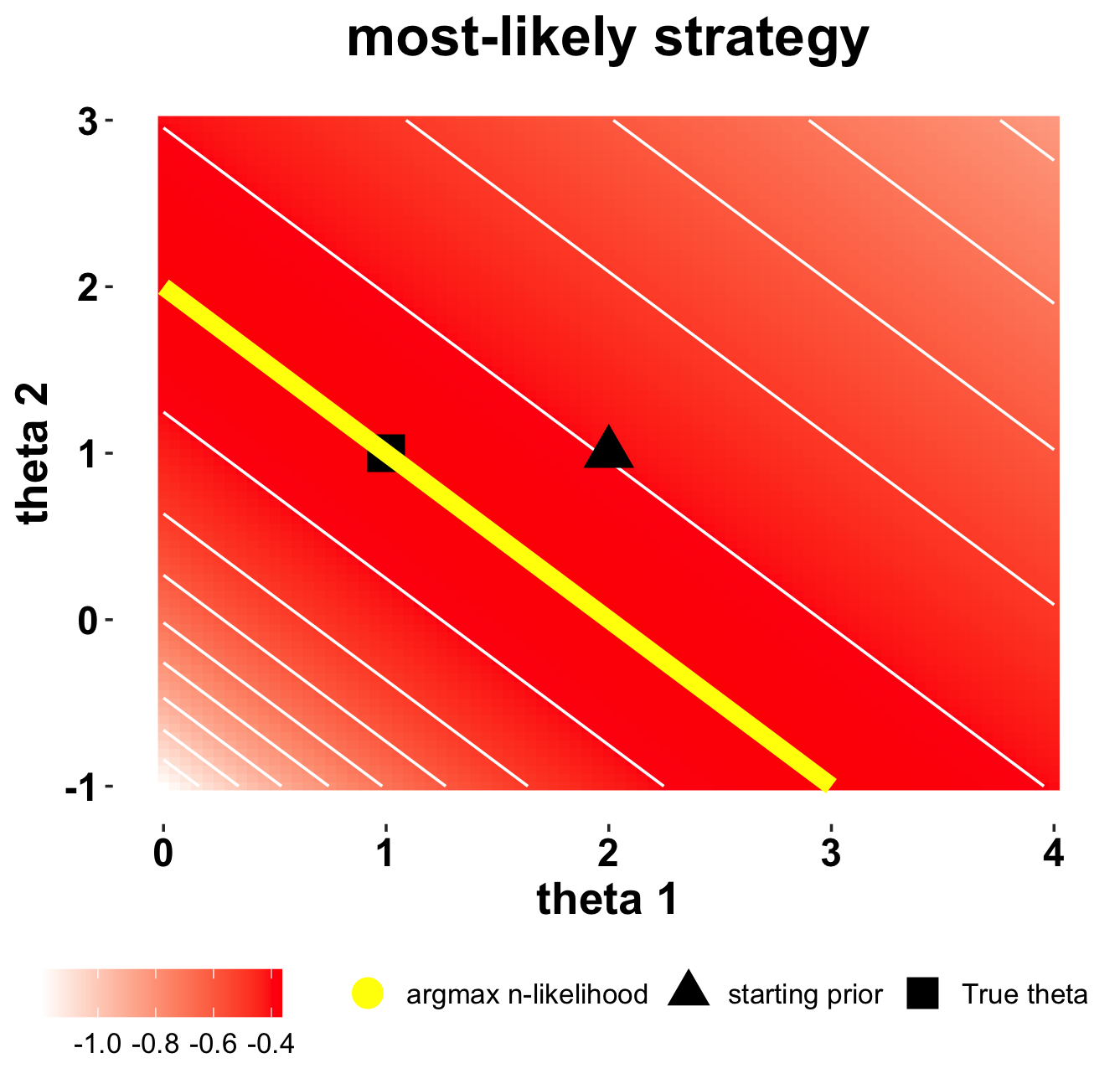}
\end{subfigure}
\begin{subfigure}{0.49\textwidth}
	\includegraphics[width=\textwidth]{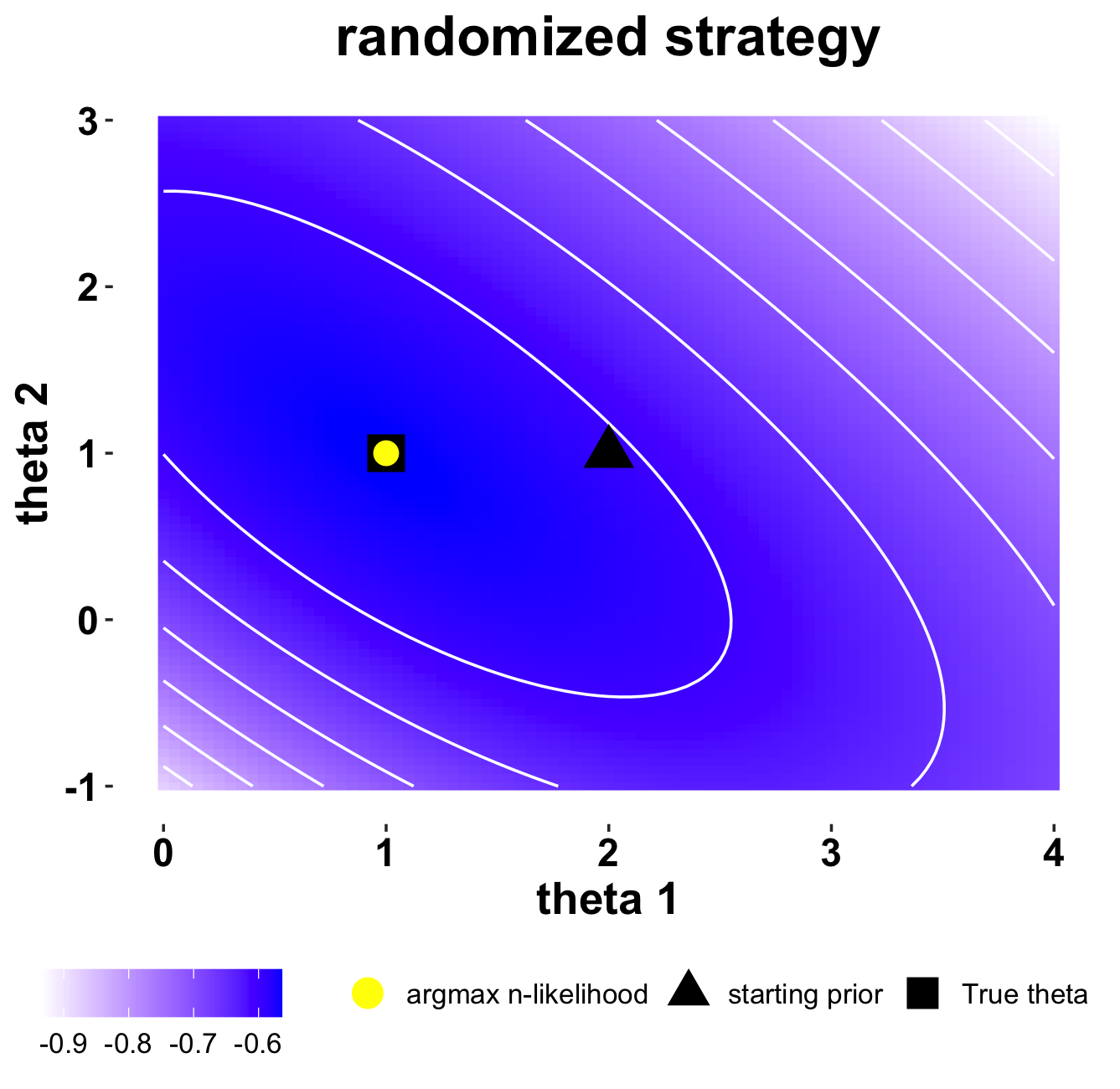}
\end{subfigure}
\caption{Estimates of the expected log-likelihood $Q_{1000}(\vartheta)$ as in (\ref{Qn_text}) for the most-likely (left) and randomized (right) strategies. Estimates are based on 50 replications. The color bar shows the level of the likelihood. For the most-likely strategy, the likelihood is degenerate: it is maximized in a segment (in yellow). In the case of the randomized strategy, the likelihood is maximized at the true $\theta$ (black square).}
\label{2Dlikelihoods}
\end{figure}

Finally, Figure \ref{2Dlikelihoods} shows the average (over realizations) level of the terminal log-likelihood $Q_{1000}(\vartheta)$ as a function of $\vartheta$ for each strategy, where 
\begin{equation}\label{Qn_text}
Q_n(\vartheta)=\frac{1}{n}\sum_{i=1}^n\log p(Y_i|X_i,\vartheta).
\end{equation}
Again the black square in each plot corresponds to $\vartheta_{true}$ and the triangle corresponds to the starting prior mean $\int_{\vartheta\in\Theta}\vartheta d\pi_0(\vartheta)$. In the case of the most-likely strategy (left), the likelihood flattens out in one direction: we do not have identification, which is due to the limit $\mathbb{E}[XX']$ (matrix of ones) being singular and $Q_{1000}(\vartheta)$ is maximized at an entire line, represented by the yellow segment. In the case of the randomized most-likely strategy (right), $Q_{1000}(\vartheta)$ is uniquely maximized at the yellow point which coincides with the true parameter value. 

\newpage
\section{Comparison with Thompson sampling}\label{section_thompson}
We now compare our proposed randomized most-likely strategy, denoted RML hereafter, to Thompson sampling (see, e.g., \citealp{agrawal2012analysis} and \citealp{kaptein2015use}) with respect to how well each method learns the fraud probability of given claim characteristics. Our simulations suggest that, when the true model is parametric, Thompson sampling can be inefficient in learning, particularly in low fraud probability regions. 

We assume a logistic-type binary regression data-generating process in one dimension for the fraud probability of any $x\in[0,1]$ given by
\begin{equation}
p(Y=1|X=x,\theta) = (1+\exp(-\theta(x^2+5x^3)))^{-1} = g(u(x)\theta)
\end{equation}
where $u(x):=x^2+5x^3$. For the RML strategy, we assume a normal distribution $\mathcal{N}(2,0.75)$ as starting prior $\pi_0$ for $\theta$. Selection is done by drawing from (\ref{prob_weights_RML}) at each step and, with every new observation, a posterior distribution for $\theta$ is obtained using Bayes' rule as in (\ref{bayes_1}). Based on the $n^{th}$ posterior for $\theta$, i.e., $\pi_n$, we define the $n^{th}$ posterior mean for the fraud probability $p(Y=1|X=x, \theta)$ by
\begin{equation}
\mathrm{E}^{\pi}_{\text{RML},n} := \int_{\vartheta\in\Theta}g(u(x)\vartheta) \pi_n(\vartheta)d\vartheta
\label{posterior_mean_fraud_prob_RML}
\end{equation}
and, similarly, the $n^{th}$ posterior variance of $p(Y=1|X=x, \theta)$ by
\begin{equation}
\mathrm{V}^{\pi}_{\text{RML},n} := \int_{\vartheta\in\Theta}g(u(x)\vartheta)^2 \pi_n(\vartheta)d\vartheta - \Big(\mathrm{E}^{\pi}_{\text{RML}}\Big)^2.
\label{posterior_variance_fraud_prob_RML}
\end{equation}
For Thompson sampling, we assume $K\geq2$ arms and partition $[0,1]$ into $K$ subsets 
\begin{equation}
[0,1] = [0,\frac{1}{K}]\cup[\frac{1}{K},\frac{2}{K}]\cup\cdots\cup[\frac{K-1}{K},1]
\end{equation}
where each segment corresponds to an arm. We assume that claims within an arm have the same fraud probability and denote by $\mu_1,\dots,\mu_K$ these probabilities. As starting prior for every $\mu_k$ we take a Beta distribution $\mathrm{Beta}(1, 1)$ which is the uniform on $[0,1]$: we start with zero knowledge on the fraud probabilities of each arm. To ``play" the bandit using Thompson sampling means drawing a realization from the prior of each arm and picking the arm with the highest draw. Then, we randomly select a claim within that arm and observe its fraud outcome. If the observed outcome is 1 (resp. 0), then the posterior for that arm is $\mathrm{Beta}(2, 1)$ (resp. $\mathrm{Beta}(1, 2)$). Noting $\alpha_{n,k} = S_{n,k} + 1$ and $\beta_{n,k} = F_{n,k} + 1$ where $S_{n,k}$ and $F_{n,k}$ are the number of observed successes and failures respectively for arm $k$ at time $n$, the $n^{th}$ posterior mean fraud probability for that arm $k$ is then given by
\begin{equation}
\mathrm{E}^{k}_{\text{THO},n} := \frac{\alpha_{n,k}}{\alpha_{n,k}+\beta_{n,k}}
\label{posterior_mean_fraud_prob_THO}
\end{equation}
and, similarly, the $n^{th}$ posterior variance by
\begin{equation}
\mathrm{V}^{k}_{\text{THO},n} := \frac{\alpha_{n,k}\ \beta_{n,k}}{(\alpha_{n,k}+\beta_{n,k})^2(\alpha_{n,k}+\beta_{n,k}+1)}.
\label{posterior_variance_fraud_prob_THO}
\end{equation}

\noindent \textbf{Simulation} \ \ We ran the above two strategies with $n=1000$ investigations where, at each step, we receive 100 claims (equally spaced observations in $[0,1]$) and choose one of them for investigation according to each strategy. For Thompson sampling, we used $K=50$ arms (hence, once an arm is selected, there are only two claims to randomly choose from). Both strategies being random, we repeated 50 realizations of each and observed the expected (sample average) posterior mean and variance of the fraud probability for each method. 

\begin{figure}[htbp]
\begin{subfigure}{0.43\textwidth}
	\includegraphics[width=\textwidth]{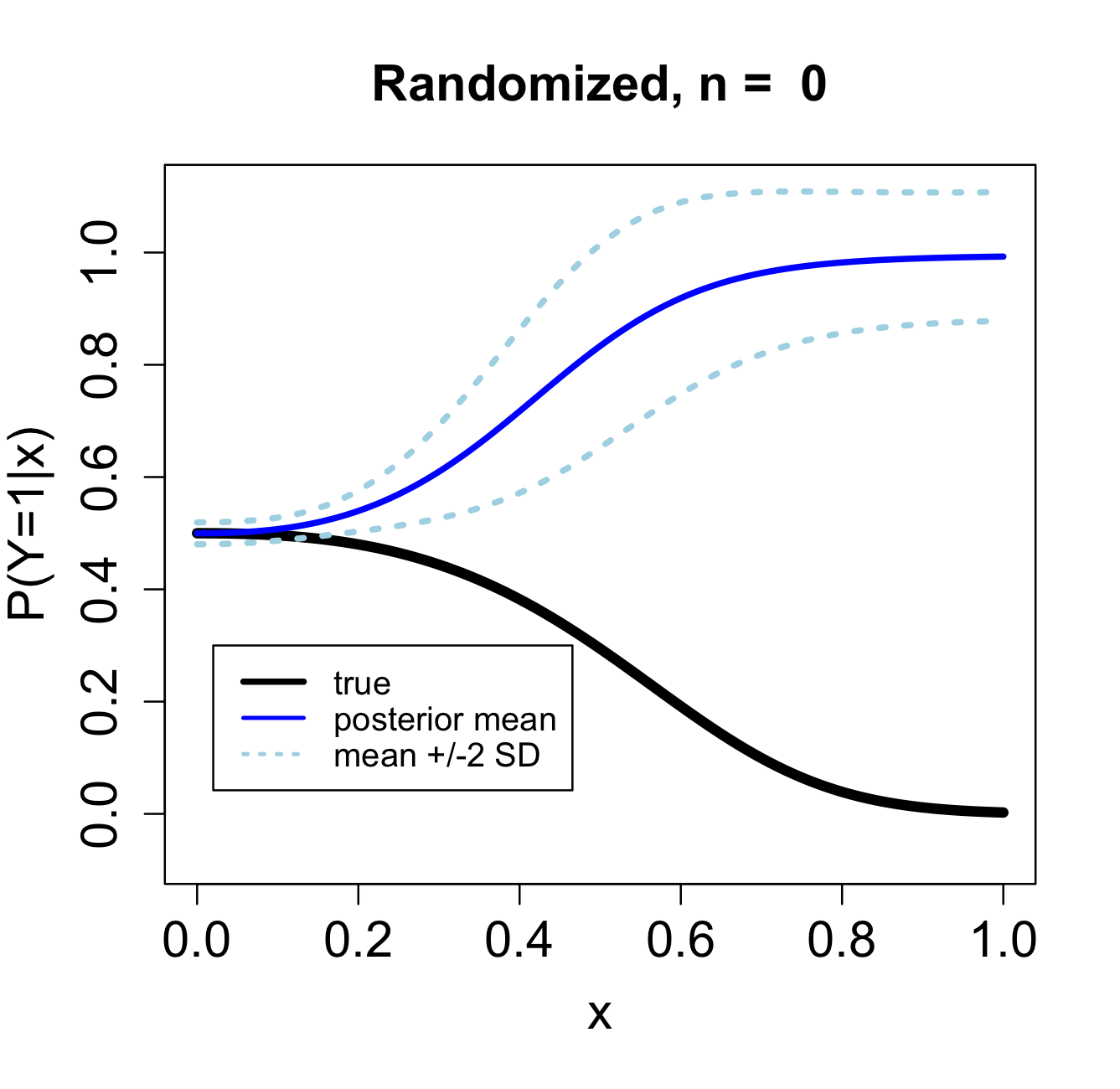}
\end{subfigure}
\begin{subfigure}{0.43\textwidth}
	\includegraphics[width=\textwidth]{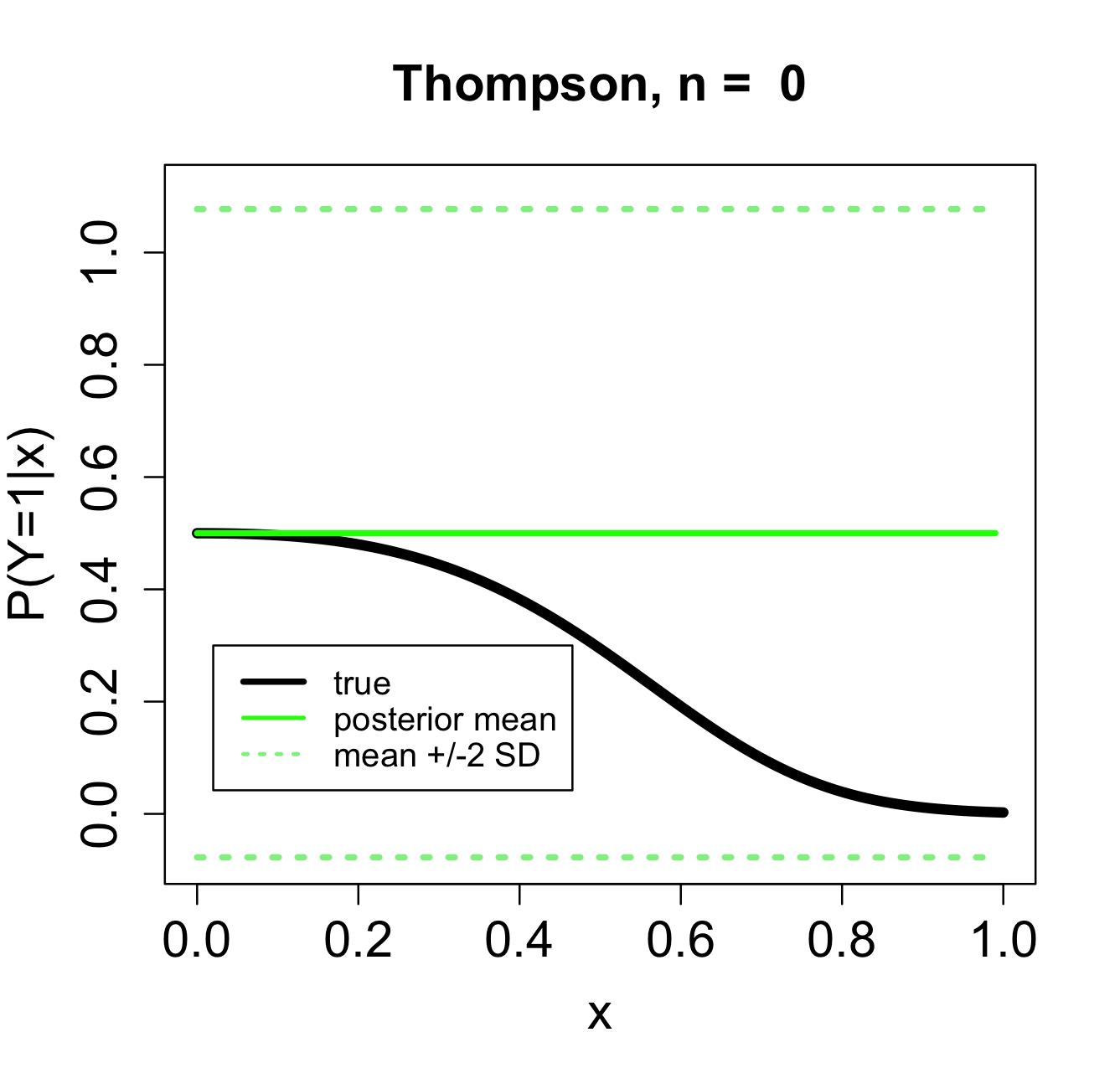}
\end{subfigure}
\begin{subfigure}{0.43\textwidth}
	\includegraphics[width=\textwidth]{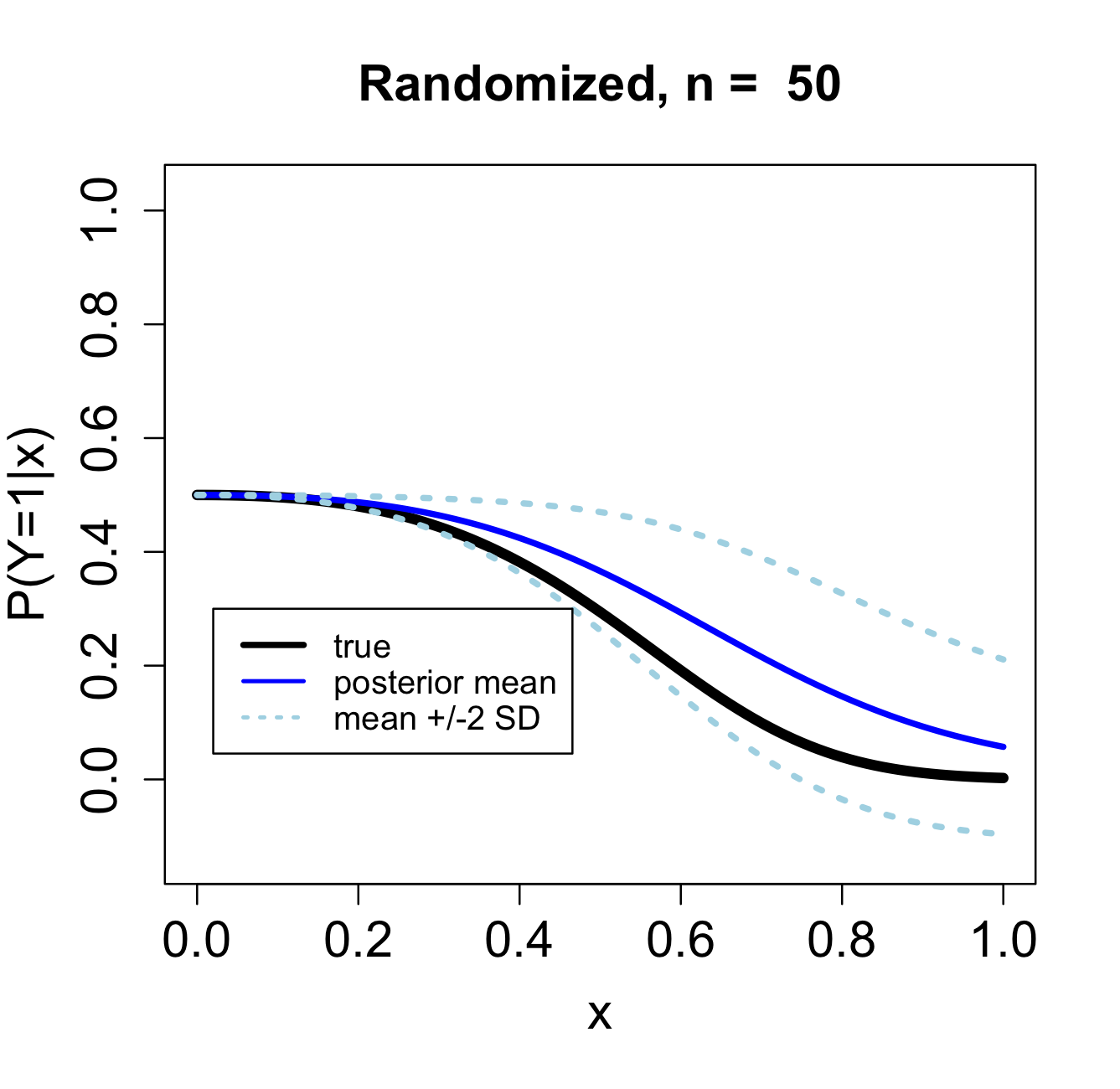}
\end{subfigure}
\begin{subfigure}{0.43\textwidth}
	\includegraphics[width=\textwidth]{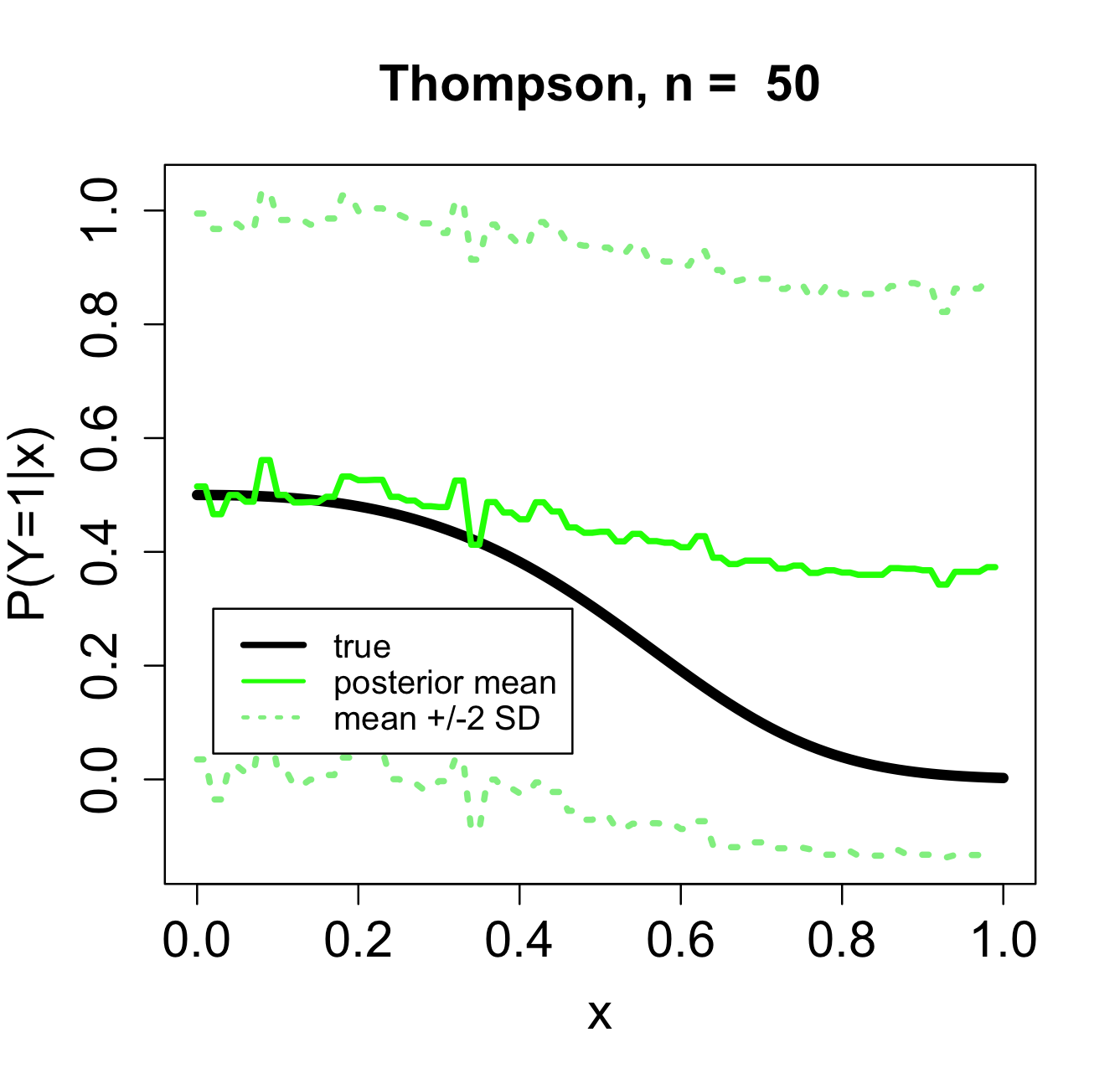}
\end{subfigure}
\begin{subfigure}{0.43\textwidth}
	\includegraphics[width=\textwidth]{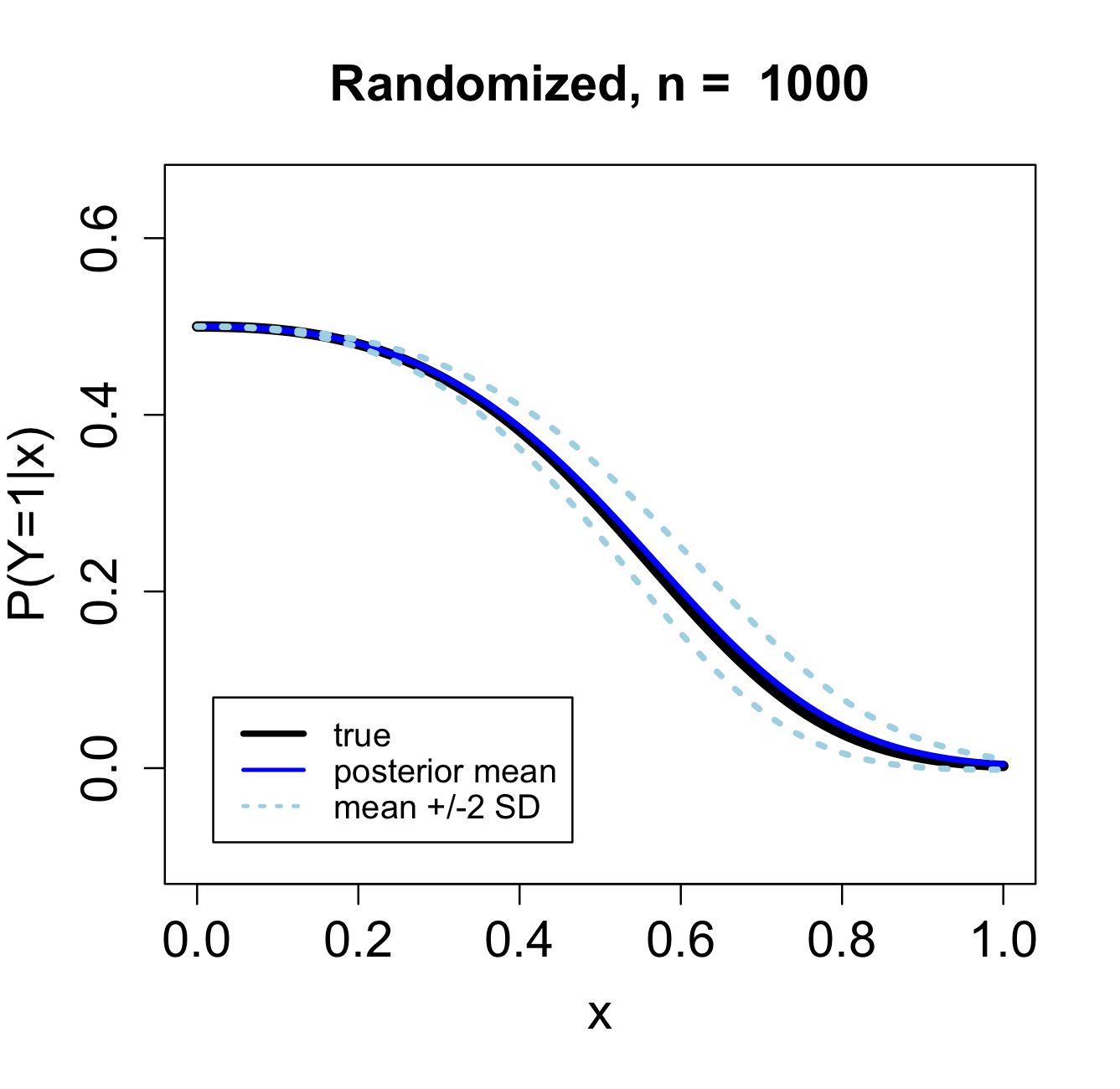}
\end{subfigure}
\begin{subfigure}{0.43\textwidth}
	\includegraphics[width=\textwidth]{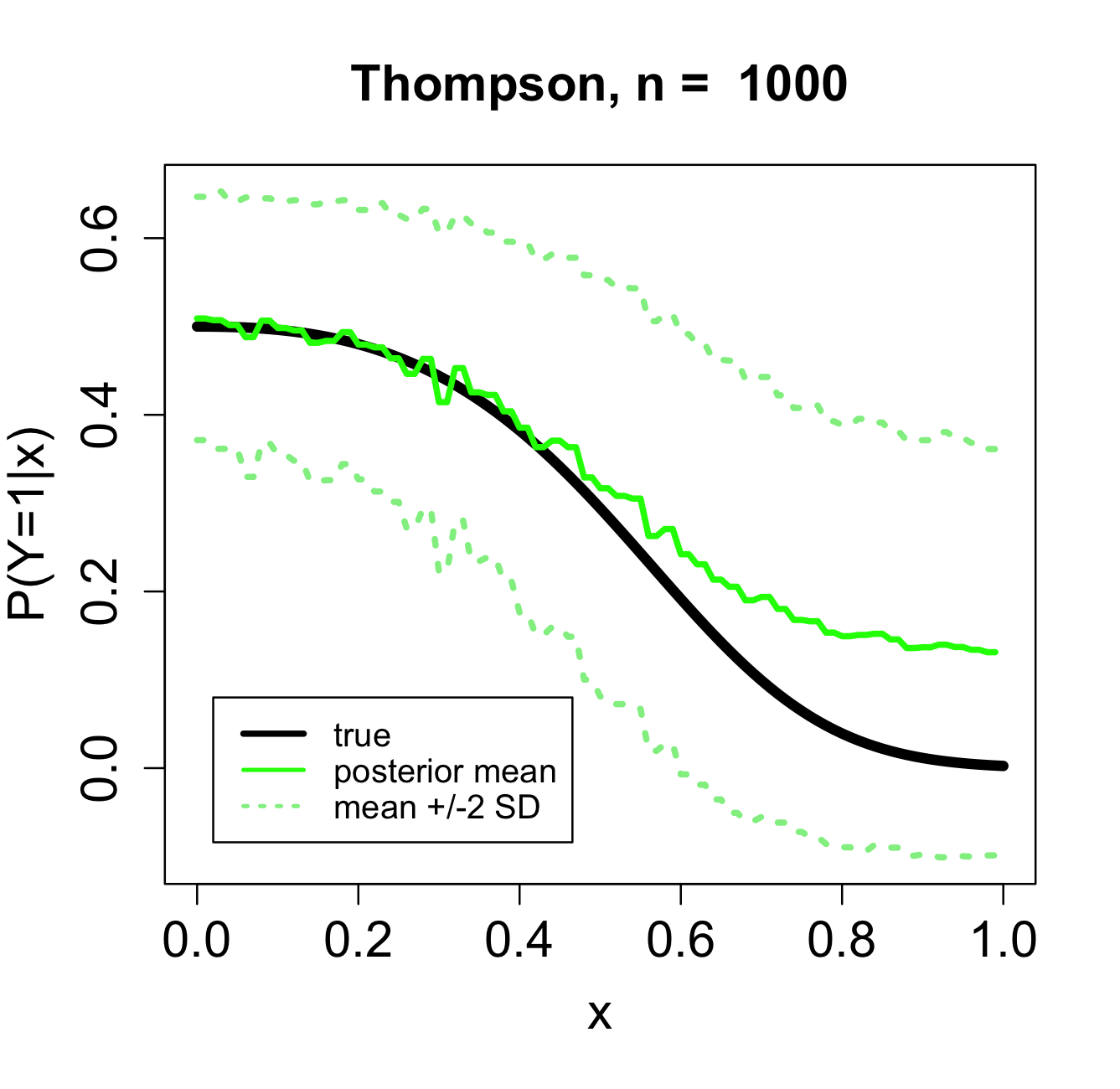}
\end{subfigure}
\caption{RML strategy (left) versus Thompson sampling (right) after $n=0$, $n=50$ and $n=1000$ claim investigations. Solid coloured lines correspond to posterior mean fraud probabilities as in (\ref{posterior_mean_fraud_prob_RML}) and (\ref{posterior_mean_fraud_prob_THO}) respectively. Dotted lines are posterior means $\pm 2 $ posterior standard deviations as in (\ref{posterior_mean_pm_2_sd}). In black are the true probabilities $g(u(x)\theta)$ given $x$. We observe that Thompson sampling is inefficient in learning low fraud probabilities.}
\label{figure_Thompson_comparison}
\end{figure}

Figure \ref{figure_Thompson_comparison} shows the obtained results after respectively 0, 50 and 1000 investigations. In each plot, the black line corresponds to the true fraud probability $g(u(x)\theta)$ given $x$ (on the horizontal axis) where the true parameter is $\theta=-1$. The coloured solid line (respectively blue for the RML strategy and green for Thompson sampling) is the average (over 50 realizations) posterior mean fraud probability (respectively $\mathrm{E}^{\pi}_{\text{RML},n}$ and $\mathrm{E}^{k}_{\text{THO},n}$). The coloured dotted lines are respectively 
\begin{equation}
\mathrm{E}^{\pi}_{\text{RML},n} \pm 2 \sqrt{\mathrm{V}^{\pi}_{\text{RML},n}} \text{\ \ and \ } \mathrm{E}^{k}_{\text{THO},n} \pm 2 \sqrt{\mathrm{V}^{k}_{\text{THO},n}}.
\label{posterior_mean_pm_2_sd}
\end{equation}
At the start, i.e., $n=0$, the belief for the RML strategy is that $\theta$ is positive, hence the opposite slope. For Thompson sampling, each arm's starting belief is uniform in $[0,1]$ hence the mean at $0.5$ for each arm. As $n$ increases, the posteriors for $\theta$ move to negative values and concentrate more and more around the true $\theta=-1$, ultimately learning the correct model. While Thompson sampling early learns the high fraud probabilities, it struggles to learn the low fraud probabilities, which persists when we increase $n$ even more. We argue that, with Thompson sampling, arms with low believed fraud probabilities are not played often, which makes this strategy less efficient when the goal is learning. Finally, even in high fraud probabilities Thompson sampling shows larger posterior variances because the more the arms, the fewer actual observations in each arm, and hence Thompson sampling is less efficient than the RML strategy.

\newpage
\section{Practical considerations and future work} \label{section_practical_guidelines}
We end this paper with a discussion on practical considerations and possible future work. There is always a trade-off between exploration and exploitation when selecting claims for investigation: some claims might be suspicious, others can be informative. The typical selection strategy used by insurers focuses on exploitation and this can lead to incorrect learning of the true fraud characteristics. Instead, we recommend incorporating some exploration. Several methods exist to do so, and a general comparison is beyond the scope of this paper. 

We have proposed a randomized alternative based on the assumption of a parametric form between claim characteristics and fraud. Proving consistency for this strategy is something we leave for future work, as the standard techniques do not trivially apply. Our simulations suggest however that this strategy correctly learns over time. It is similar to Thompson sampling in the sense that selection is proportional to the believed fraud probabilities. A disadvantage of MAB strategies is non-interpretability, contrary to our binary regression framework, where, e.g., in a logit model, the coefficients can be explained. 

Our proposed strategy can be more generally applied to other supervised prediction methods, as long as they yield estimates for the fraud probability of incoming claims. Suppose for example that a random forest (\citealp{breiman2001random}) is trained and that 10 car accident claims are received on a given day. Based on the forest predictions, these 10 claims can be ordered. Suppose moreover that only one claim can be investigated that day. The typical choice would be to investigate the claim at the top of the list. Our proposed strategy, instead, would pick the same claim with a probability that is proportional to the predictions, hence allowing for some exploration. 
An interesting next step would be to compare the typical selection strategy with its randomized analogue for a variety of prediction methods. Empirically comparing methods in real datasets, however, would be more challenging because for that we would need a sample of claims that have been randomly chosen for investigation and fraud has been established, i.e., a sample without selection bias.


\section{Conclusion}\label{section_conclusion}
In this paper we have formalized selection as a map from posterior beliefs to distributions on the covariate space. Under the assumption that selection does not depend on the parameter of interest, we have shown that model updating and maximum-likelihood estimation can be implemented as if the collected data was iid. We have defined consistency of selection strategies and conjectured sufficient conditions for consistency. We have shown simulation results suggesting that the typical claim selection strategy used by insurers can lead to inconsistent learning and proposed a randomized alternative that, in simulations, converges to the true model. Moreover, our simulations suggest that Thompson sampling can be less efficient than our proposed strategy when the data-generating process is parametric, particularly in regions of low fraud probability.


\newpage
\appendix
\section{Proofs}\label{section_appendix}

\subsection{Proof of Proposition \ref{proposition_bayes}} \label{appendix_bayes}
Bayes' rule implies
\begin{equation}
\pi_{i+1}
=\frac{p(x_{i+1},y_{i+1}|x_1,y_1,\dots,x_i,y_i,\vartheta)\pi_i}{p(x_{i+1},y_{i+1}|x_1,y_1,\dots,x_i,y_i)}.
\label{appendix_eq_needed_1}
\end{equation}
Using (\ref{dgp_assumption}) and (\ref{selection_assumption}) we have
\begin{align}
& \ p(x_{i+1},y_{i+1}|x_1,y_1,\dots,x_i,y_i,\vartheta) \nonumber \\
= & \ p(x_{i+1}|x_1,y_1,\dots,x_i,y_i,\vartheta) \ p(y_{i+1}|x_1,y_1,\dots,x_i,y_i,x_{i+1},\vartheta) \\
= & \ p(x_{i+1}|x_1,y_1,\dots,x_i,y_i) \ p(y_{i+1}|x_{i+1}, \vartheta). \nonumber
\end{align}
Similarly, 
\begin{align}
& p(x_{i+1},y_{i+1}|x_1,y_1,\dots,x_i,y_i) \nonumber \\
= & \int_{\vartheta\in\Theta} p(x_{i+1},y_{i+1}|x_1,y_1,\dots,x_i,y_i,\vartheta) \pi_i(\vartheta)d\vartheta  \\
= & \int_{\vartheta\in\Theta} p(x_{i+1}|x_1,y_1,\dots,x_i,y_i) \ p(y_{i+1}|x_{i+1}, \vartheta) \pi_i(\vartheta)d\vartheta \text{ \ using (\ref{dgp_assumption}) and (\ref{selection_assumption})} \nonumber \\
= & \ p(x_{i+1}|x_1,y_1,\dots,x_i,y_i) \int_{\vartheta\in\Theta} \ p(y_{i+1}|x_{i+1}, \vartheta) \pi_i(\vartheta)d\vartheta \nonumber \\
= & \ p(x_{i+1}|x_1,y_1,\dots,x_i,y_i) \ p(y_{i+1}|x_{i+1}). \nonumber
\end{align}
Finally, the term $p(x_{i+1}|x_1,y_1,\dots,x_i,y_i)$ simplifies in the numerator and denominator of (\ref{appendix_eq_needed_1}), which concludes the proof. In the particular case $p(x_{i+1}|x_1,y_1,\dots,x_i,y_i)=p(x_{i+1}|\pi_i)$ the same proof holds.

\subsection{Proof of Theorem \ref{theorem_consistency}} \label{appendix_consistency_thm}

We proceed in two steps. First, from martingale theory, we obtain that the sequence of posterior means converges. Second, from the recovery condition (\ref{recovery_condition}) we obtain that the limit is the one that we want.

Let $\mathcal{A}$ and $\mathcal{B}$ be the canonical $\sigma$-algebras associated with $\mathbb{R}^k$ and $\{0,1\}$ respectively. Let $F_n$ be the $\sigma$-algebra generated by $(X_1,Y_1),\dots,(X_n,Y_n)$. The sequence $\{F_n\}_{n\geq1}$ forms a filtration of $\big((\mathbb{R}^k\times\{0,1\})^\infty,(\mathcal{A}\otimes\mathcal{B})^{\otimes\infty}\big)$ where $(\mathcal{A}\otimes\mathcal{B})^{\otimes\infty}$ is the infinite product $\sigma$-algebra. Denote by $F_\infty$ the filtration's limit. 

Let $h:\Theta\longrightarrow\mathbb{R}$ be a continuous and bounded function and define
\begin{equation}
M_n:=
\int_{\vartheta\in\Theta}h(\vartheta)\pi_n(\vartheta) d\vartheta
\end{equation}
for $n\geq1$. These integrals are well defined due to the boundness of $h$ and the integrability of $\pi_n$. By construction, we can interpret $M_n$ as the expectation of $h(\theta)$ conditionally on the filtration $F_n$, i.e.,
\begin{equation}
M_n = \mathrm{E}[h(\theta)|F_n]
\end{equation}
where the expectation is taken with respect to the posterior $\pi_n$. Hence, the sequence $M_n$ forms a martingale with respect to $F_n$. Then, from Lévy's zero–one law 
the sequence $\{M_n\}_{n\geq1}$ converges almost surely to $\mathbb{E}[h(\theta)|F_\infty]$. From the recovery condition (\ref{recovery_condition}), $\theta$ is $F_{\infty}$-measurable and, thus, $\mathbb{E}[h(\theta)|F_\infty] = h(\theta)$ almost surely, which concludes the proof.

\subsection{Proof of Proposition \ref{proposition_MLE}}\label{appendix_proposition_likelihood}
The maximum likelihood estimator is defined by
\begin{equation}
\hat{\vartheta}_n := \argmax_{\vartheta\in\Theta} p(X_1,Y_1,\dots,X_n,Y_n|\vartheta). 
\end{equation}
We have
\begin{align}
& p(X_1,Y_1,\dots,X_n,Y_n|\vartheta) \\
= & \prod_{i=1}^np(X_i,Y_i|X_1,Y_1,\dots,X_{i-1},Y_{i-1},\vartheta) \nonumber \text{ \ where the first term is $p(X_1,Y_1|\vartheta)$} \\
= & \prod_{i=1}^np(X_i|X_1,Y_1,\dots,X_{i-1},Y_{i-1}) p(Y_i|X_i,\vartheta) \text{ \ using (\ref{dgp_assumption}) and (\ref{selection_assumption}).}  \nonumber 
\end{align}
Taking the $\log$ does not change the maximum. Hence 
\begin{equation}
\hat{\vartheta}_n = \argmax_{\vartheta\in\Theta} \Big\{ \sum_{i=1}^n \log p(X_i|X_1,Y_1,\dots,X_{i-1},Y_{i-1}) + \sum_{i=1}^n \log p(Y_i|X_i,\vartheta) \Big\} 
\label{equation_ref_appendix}
\end{equation}
The first sum in (\ref{equation_ref_appendix}) does not depend on $\vartheta$, hence
\begin{equation}
\hat{\vartheta}_n = \argmax_{\vartheta\in\Theta}  \sum_{i=1}^n \log p(Y_i|X_i,\vartheta) 
\label{equation_ref_appendix_2}
\end{equation}
which concludes the proof.

\subsection{Towards the proof of Conjecture \ref{conjecture_empirical}}\label{appendix_conjecture}

Note the normalized log-likelihood
\begin{equation}\label{Qn}
Q_n(\vartheta):=\frac{1}{n}\sum_{i=1}^n\log p(Y_i|X_i,\vartheta).
\end{equation}
Then $\hat{\vartheta}_n=\argmax_{\vartheta\in\Theta}Q_n(\vartheta)$. 

Let $\vartheta_0$ be a realization of $\theta$. Suppose that the sequence $\{\hat{\vartheta}_n\}_{n\geq1}$ is strongly consistent for $\vartheta_0$ under actual sampling, i.e. with selection. That is $\hat{\vartheta}_n\rightarrow\vartheta_0$ almost surely. 
Define $Q_0: (\mathbb{R}^k\times\{0,1\})^\infty\rightarrow \Theta$ almost surely by 
\begin{equation}
Q_0\big((X_1,Y_1),\dots,(X_\infty,Y_\infty)\big):=\lim_{n\rightarrow\infty}\hat{\vartheta}_n.
\end{equation}
By the strong consistency assumption, we have $Q_0\big((X_1,Y_1),\dots,(X_\infty,Y_\infty)\big)=\vartheta_0$ almost surely. This being true for every realization of $\theta$, the recovery condition (\ref{recovery_condition}) is then satisfied as long as the function $Q_0$ is measurable, which is guaranteed by the continuity of $g$. Hence, showing strong consistency of the maximum likelihood estimator implies the recovery condition and, hence, consistency of the selection strategy.



In order to establish consistency of the MLE, we follow, e.g., \citet{newey1994large} [Theorem 2.1, pp 2121]. $\{\hat{\vartheta}_n\}$ is strongly consistent if there exists a function $Q_0$ of $\vartheta$ such that 
\begin{enumerate}[label=(\roman*)]
\item $Q_0$ is continuous,
\item $Q_0$ is uniquely maximized at $\vartheta_0$, and
\item $Q_n$ defined in (\ref{Qn}) converges uniformly a.s. to $Q_0$.
\end{enumerate}
Let 
\begin{equation}
Q_0(\vartheta):=\mathrm{E}_{\vartheta_0}[\log p(Y|X,\vartheta)|X]
\end{equation}
where the expectation is taken under (\ref{model}) with $\theta=\vartheta_0$. $X$ is a column vector such that $\lim_{n\rightarrow\infty}\frac{1}{n}\mathbf{X}'\mathbf{X}=\mathbb{E}[XX']$. Note that $\mathbf{X}'\mathbf{X}$ is symmetric and positive semi definite, hence invertibility is equivalent to positive definite. We have 
\begin{equation}
Q_0(\vartheta)=g(X\vartheta_0)\log g(X\vartheta_0)+(1-g(X\vartheta_0))\log(1-g(X\vartheta_0)).
\end{equation}
From the continuity of $g$ we obtain (i). Showing (ii) is equivalent to showing identification of $\vartheta_0$ which, in turn, is obtained from the strict monotonicity of $g$ and the non-singularity of the second-moment matrix. 

The difficult part is to show (iii) under selection sampling. We conjecture that Equation (\ref{dgp_assumption}) allows us to use a uniform law of large numbers to obtain (iii). First, condition (iii) requires that observations obtained from a selection strategy are sufficiently independent for a LLN to work. Second, we believe that for the limit of $\frac{1}{n}\mathbf{X}'\mathbf{X}$ to exist and be deterministic also needs sufficiently independent observations.


\newpage
\vskip 0.2in
\bibliography{biblio}

\begin{thebibliography}{35}
\providecommand{\natexlab}[1]{#1}
\providecommand{\url}[1]{\texttt{#1}}
\expandafter\ifx\csname urlstyle\endcsname\relax
  \providecommand{\doi}[1]{doi: #1}\else
  \providecommand{\doi}{doi: \begingroup \urlstyle{rm}\Url}\fi

\bibitem[Agrawal and Goyal(2012)]{agrawal2012analysis}
Shipra Agrawal and Navin Goyal.
\newblock Analysis of thompson sampling for the multi-armed bandit problem.
\newblock In \emph{Conference on learning theory}, pages 39--1. JMLR Workshop
  and Conference Proceedings, 2012.

\bibitem[Auer et~al.(2002)Auer, Cesa-Bianchi, and Fischer]{auer2002finite}
Peter Auer, Nicolo Cesa-Bianchi, and Paul Fischer.
\newblock Finite-time analysis of the multiarmed bandit problem.
\newblock \emph{Machine learning}, 47:\penalty0 235--256, 2002.

\bibitem[Baesens(2023)]{baesens2023fraud}
Bart Baesens.
\newblock Fraud analytics: a research.
\newblock \emph{Journal of Chinese Economic and Business Studies}, 21\penalty0
  (1):\penalty0 137--141, 2023.

\bibitem[Baesens et~al.(2021{\natexlab{a}})Baesens, H{\"o}ppner, Ortner, and
  Verdonck]{baesens2021robrose}
Bart Baesens, Sebastiaan H{\"o}ppner, Irene Ortner, and Tim Verdonck.
\newblock robrose: A robust approach for dealing with imbalanced data in fraud
  detection.
\newblock \emph{Statistical Methods \& Applications}, 30\penalty0 (3):\penalty0
  841--861, 2021{\natexlab{a}}.

\bibitem[Baesens et~al.(2021{\natexlab{b}})Baesens, H{\"o}ppner, and
  Verdonck]{baesens2021data}
Bart Baesens, Sebastiaan H{\"o}ppner, and Tim Verdonck.
\newblock Data engineering for fraud detection.
\newblock \emph{Decision Support Systems}, 150:\penalty0 113492,
  2021{\natexlab{b}}.

\bibitem[Barton et~al.(2024)Barton, Burnett, Gunny, and
  Miller]{barton2024importance}
F~Jane Barton, Brian~M Burnett, Katherine Gunny, and Brian~P Miller.
\newblock The importance of separating the probability of committing and
  detecting misstatements in the restatement setting.
\newblock \emph{Management Science}, 70\penalty0 (1):\penalty0 32--53, 2024.

\bibitem[Baumann(2021)]{baumann2021improving}
Michaela Baumann.
\newblock Improving a rule-based fraud detection system with classification
  based on association rule mining.
\newblock In \emph{INFORMATIK 2021}, pages 1121--1134. Gesellschaft f{\"u}r
  Informatik, Bonn, 2021.

\bibitem[Benedek et~al.(2022)Benedek, Ciumas, and Nagy]{benedek2022automobile}
Botond Benedek, Cristina Ciumas, and B{\'a}lint~Zsolt Nagy.
\newblock Automobile insurance fraud detection in the age of big data--a
  systematic and comprehensive literature review.
\newblock \emph{Journal of Financial Regulation and Compliance}, 30\penalty0
  (4):\penalty0 503--523, 2022.

\bibitem[Bouneffouf and Rish(2019)]{bouneffouf2019survey}
Djallel Bouneffouf and Irina Rish.
\newblock A survey on practical applications of multi-armed and contextual
  bandits.
\newblock \emph{\href{https://arxiv.org/abs/1904.10040}{arXiv preprint
  arXiv:1904.10040}}, 2019.

\bibitem[Breiman(2001)]{breiman2001random}
Leo Breiman.
\newblock Random forests.
\newblock \emph{Machine learning}, 45:\penalty0 5--32, 2001.

\bibitem[Caudill et~al.(2005)Caudill, Ayuso, and Guill{\'e}n]{caudill2005fraud}
Steven~B Caudill, Mercedes Ayuso, and Montserrat Guill{\'e}n.
\newblock Fraud detection using a multinomial logit model with missing
  information.
\newblock \emph{Journal of Risk and Insurance}, 72\penalty0 (4):\penalty0
  539--550, 2005.

\bibitem[Cecchini et~al.(2010)Cecchini, Aytug, Koehler, and
  Pathak]{cecchini2010detecting}
Mark Cecchini, Haldun Aytug, Gary~J Koehler, and Praveen Pathak.
\newblock Detecting management fraud in public companies.
\newblock \emph{Management Science}, 56\penalty0 (7):\penalty0 1146--1160,
  2010.

\bibitem[Collier and Llorens(2018)]{collier2018deep}
Mark Collier and Hector~Urdiales Llorens.
\newblock Deep contextual multi-armed bandits.
\newblock \emph{\href{https://arxiv.org/abs/1807.09809}{arXiv preprint
  arXiv:1807.09809}}, 2018.

\bibitem[Cong et~al.(2023)Cong, Li, Tang, and Yang]{cong2023crypto}
Lin~William Cong, Xi~Li, Ke~Tang, and Yang Yang.
\newblock Crypto wash trading.
\newblock \emph{Management Science}, 69\penalty0 (11):\penalty0 6427--6454,
  2023.

\bibitem[Dimri et~al.(2022)Dimri, Paul, Girish, Lee, Afra, and
  Jakubowski]{dimri2022multi}
Anuj Dimri, Arindam Paul, Deeptha Girish, Peng Lee, Sardar Afra, and Andrew
  Jakubowski.
\newblock A multi-input multi-label claims channeling system using
  insurance-based language models.
\newblock \emph{Expert Systems with Applications}, 202:\penalty0 117166, 2022.

\bibitem[Doob(1949)]{doob1949application}
Joseph~L Doob.
\newblock Application of the theory of martingales.
\newblock \emph{Le calcul des probabilites et ses applications}, pages 23--27,
  1949.

\bibitem[H{\"o}ppner et~al.(2022)H{\"o}ppner, Baesens, Verbeke, and
  Verdonck]{hoppner2022instance}
Sebastiaan H{\"o}ppner, Bart Baesens, Wouter Verbeke, and Tim Verdonck.
\newblock Instance-dependent cost-sensitive learning for detecting transfer
  fraud.
\newblock \emph{European Journal of Operational Research}, 297\penalty0
  (1):\penalty0 291--300, 2022.

\bibitem[Jung et~al.(2012)Jung, Martin, Ernst, and Leduc]{jung2012contextual}
Tobias Jung, Sylvain Martin, Damien Ernst, and Guy Leduc.
\newblock Contextual multi-armed bandits for the prevention of spam in voip
  networks.
\newblock \emph{\href{https://arxiv.org/abs/1201.6181}{arXiv preprint
  arXiv:1201.6181}}, 2012.

\bibitem[Kaptein(2015)]{kaptein2015use}
Maurits Kaptein.
\newblock The use of thompson sampling to increase estimation precision.
\newblock \emph{Behavior research methods}, 47\penalty0 (2):\penalty0 409--423,
  2015.

\bibitem[Lessmann et~al.(2008)Lessmann, Baesens, Mues, and
  Pietsch]{lessmann2008benchmarking}
Stefan Lessmann, Bart Baesens, Christophe Mues, and Swantje Pietsch.
\newblock Benchmarking classification models for software defect prediction: A
  proposed framework and novel findings.
\newblock \emph{IEEE transactions on software engineering}, 34\penalty0
  (4):\penalty0 485--496, 2008.

\bibitem[Liu et~al.(2020)Liu, Yang, Xu, Derrick, Stubbs, and
  Stockdale]{liu2020automobile}
Xi~Liu, Jian-Bo Yang, Dong-Ling Xu, Karim Derrick, Chris Stubbs, and Martin
  Stockdale.
\newblock Automobile insurance fraud detection using the evidential reasoning
  approach and data-driven inferential modelling.
\newblock In \emph{2020 IEEE International Conference on Fuzzy Systems
  (FUZZ-IEEE)}, pages 1--7. IEEE, 2020.

\bibitem[Lu et~al.(2006)Lu, Boritz, and Covvey]{lu2006adaptive}
Fletcher Lu, J~Efrim Boritz, and Dominic Covvey.
\newblock Adaptive fraud detection using benford’s law.
\newblock In \emph{Advances in Artificial Intelligence: 19th Conference of the
  Canadian Society for Computational Studies of Intelligence, Canadian AI 2006,
  Qu{\'e}bec City, Qu{\'e}bec, Canada, June 7-9, 2006. Proceedings 19}, pages
  347--358. Springer, 2006.

\bibitem[Miller(2018)]{miller2018detailed}
Jeffrey~W Miller.
\newblock A detailed treatment of doob's theorem.
\newblock \emph{\href{https://arxiv.org/abs/1801.03122}{arXiv preprint
  arXiv:1801.03122}}, 2018.

\bibitem[Newey and McFadden(1994)]{newey1994large}
Whitney~K Newey and Daniel McFadden.
\newblock Large sample estimation and hypothesis testing.
\newblock \emph{Handbook of econometrics}, 4:\penalty0 2111--2245, 1994.

\bibitem[{\'O}skarsd{\'o}ttir et~al.(2022){\'O}skarsd{\'o}ttir, Ahmed, Antonio,
  Baesens, Dendievel, Donas, and Reynkens]{oskarsdottir2022social}
Mar{\'\i}a {\'O}skarsd{\'o}ttir, Waqas Ahmed, Katrien Antonio, Bart Baesens,
  R{\'e}mi Dendievel, Tom Donas, and Tom Reynkens.
\newblock Social network analytics for supervised fraud detection in insurance.
\newblock \emph{Risk Analysis}, 42\penalty0 (8):\penalty0 1872--1890, 2022.

\bibitem[Perchet and Rigollet(2013)]{10.1214/13-AOS1101}
Vianney Perchet and Philippe Rigollet.
\newblock {The multi-armed bandit problem with covariates}.
\newblock \emph{The Annals of Statistics}, 41\penalty0 (2):\penalty0 693 --
  721, 2013.

\bibitem[Pinquet et~al.(2007)Pinquet, Ayuso, and
  Guill{\'e}n]{pinquet2007selection}
Jean Pinquet, Mercedes Ayuso, and Montserrat Guill{\'e}n.
\newblock Selection bias and auditing policies for insurance claims.
\newblock \emph{Journal of Risk and Insurance}, 74\penalty0 (2):\penalty0
  425--440, 2007.

\bibitem[Schrijver et~al.(2024)Schrijver, Sarmah, and
  El-Hajj]{schrijver2024automobile}
Gilian Schrijver, Dipti~K Sarmah, and Mohammed El-Hajj.
\newblock Automobile insurance fraud detection using data mining: A systematic
  literature review.
\newblock \emph{Intelligent Systems with Applications}, 21:\penalty0 200340,
  2024.

\bibitem[Snorovikhina and Zaytsev(2021)]{snorovikhina2021unsupervised}
Victoria Snorovikhina and Alexey Zaytsev.
\newblock Unsupervised anomaly detection for discrete sequence healthcare data.
\newblock In \emph{Analysis of Images, Social Networks and Texts: 9th
  International Conference, AIST 2020, Skolkovo, Moscow, Russia, October
  15--16, 2020, Revised Selected Papers 9}, pages 391--403. Springer, 2021.

\bibitem[Soemers et~al.(2018)Soemers, Brys, Driessens, Winands, and
  Now{\'e}]{soemers2018adapting}
Dennis Soemers, Tim Brys, Kurt Driessens, Mark Winands, and Ann Now{\'e}.
\newblock Adapting to concept drift in credit card transaction data streams
  using contextual bandits and decision trees.
\newblock \emph{Proceedings of the AAAI conference on artificial intelligence},
  32\penalty0 (1), 2018.

\bibitem[Stripling et~al.(2018)Stripling, Baesens, Chizi, and vanden
  Broucke]{stripling2018isolation}
Eugen Stripling, Bart Baesens, Barak Chizi, and Seppe vanden Broucke.
\newblock Isolation-based conditional anomaly detection on mixed-attribute data
  to uncover workers' compensation fraud.
\newblock \emph{Decision Support Systems}, 111:\penalty0 13--26, 2018.

\bibitem[Thompson(1933)]{thompson1933likelihood}
William~R Thompson.
\newblock On the likelihood that one unknown probability exceeds another in
  view of the evidence of two samples.
\newblock \emph{Biometrika}, 25\penalty0 (3/4):\penalty0 285--294, 1933.

\bibitem[Van~der Vaart(2000)]{van2000asymptotic}
Aad~W Van~der Vaart.
\newblock \emph{Asymptotic statistics}, volume~3.
\newblock Cambridge university press, 2000.

\bibitem[Van~Vlasselaer et~al.(2017)Van~Vlasselaer, Eliassi-Rad, Akoglu,
  Snoeck, and Baesens]{van2017gotcha}
V{\'e}ronique Van~Vlasselaer, Tina Eliassi-Rad, Leman Akoglu, Monique Snoeck,
  and Bart Baesens.
\newblock Gotcha! network-based fraud detection for social security fraud.
\newblock \emph{Management Science}, 63\penalty0 (9):\penalty0 3090--3110,
  2017.

\bibitem[Viaene et~al.(2005)Viaene, Dedene, and Derrig]{viaene2005auto}
Stijn Viaene, Guido Dedene, and Richard~A Derrig.
\newblock Auto claim fraud detection using bayesian learning neural networks.
\newblock \emph{Expert systems with applications}, 29\penalty0 (3):\penalty0
  653--666, 2005.

\end{thebibliography}

\end{document}